%% file: arxiv_draft.tex
\documentclass[11pt]{article}
\pdfoutput=1
\usepackage{booktabs}
\usepackage{pgfplots}
\usepackage{graphicx}
\usepackage{tikz}
\usepackage{amsmath, amsthm}
\usepackage{amssymb}
\usepackage{graphicx}
\usepackage{setspace}
\usepackage{xfrac}
\usepackage{hyperref}
\usepackage{xstring}
\usepackage{xspace}
\usepackage{bm}
\usepackage{microtype}
\usepackage{graphicx}
\usepackage{booktabs} 
\usepackage{comment}
\pgfplotsset{compat=1.14}
\usepackage{gensymb}
\usepackage{hyperref}       
\usepackage{amsfonts}       
\usepackage{nicefrac}       
\usepackage{microtype}      
\usepackage{amsmath}
\usepackage{amsfonts}
\usepackage{amssymb}
\usepackage{comment}

\usepackage{amsthm}
\usepackage{amscd}
\usepackage{t1enc}
\usepackage{enumerate}
\usepackage[mathscr]{eucal}
\usepackage{indentfirst}
\usepackage{listings}
\usepackage{graphicx}
\usepackage{graphics}
\usepackage{pict2e}
\usepackage{epic}

\usepackage{algorithm}
\usepackage{algorithmic}
\usepackage{subfig}
\usepackage{amsmath, amssymb, amscd, amsthm, amsfonts}
\usepackage{bbm}
\usepackage{booktabs}
\usepackage[shortlabels]{enumitem}
\usepackage[draft]{fixme}
\fxsetup{inline,nomargin,theme=color}

\newtheorem{assumption_supplement}{Assumption}

\usepackage{epstopdf} 
\usepackage{wrapfig}

\renewcommand{\tilde}{\widetilde}
\renewcommand{\nu}{\vartheta}

\usepackage{float}
\allowdisplaybreaks
\usepackage{fullpage}
\newtheorem{theorem}{Theorem}
\newtheorem*{theorem*}{Theorem}

\newtheorem{assumption}{Assumption}

\newtheorem*{proposition*}{Proposition}
\theoremstyle{definition}

\newtheorem*{remark*}{Remark}
\date{}
\title{\textbf{Clustering Interval-Censored Time-Series\\for Disease Phenotyping}}
\author{Irene Y. Chen \thanks{Massachusetts Institute of Technology. Email: \texttt{iychen@csail.mit.edu}} \and Rahul G. Krishnan \thanks{University of Toronto. Work done partially while at Microsoft Research New England. Email: \texttt{rahulgk@cs.toronto.edu}} \and David Sontag \thanks{Massachusetts Institute of Technology. Email: \texttt{dsontag@csail.mit.edu}}}

\begin{document}
\maketitle
\begin{abstract}
    Unsupervised learning is often used to uncover clusters in data. However, different kinds of noise may impede the discovery of useful patterns from real-world time-series data. In this work, we focus on mitigating the interference of interval censoring in the task of clustering for disease phenotyping. We develop a deep generative, continuous-time model of time-series data that clusters time-series while correcting for censorship time. We provide conditions under which clusters and the amount of delayed entry may be identified from data under a noiseless model. 
On synthetic data, we demonstrate accurate, stable, and interpretable results that outperform several benchmarks. On real-world clinical datasets of heart failure and Parkinson's disease patients, we study how interval censoring can adversely affect the task of disease phenotyping. Our model corrects for this source of error and recovers known clinical subtypes.
\end{abstract}
\section{Introduction}
Cluster analysis of time-series data is a task of interest across a variety of scientific disciplines including biology~\cite{listgarten2007bayesian}, meteorology~\cite{camargo2007cluster}, and astrophysics~\cite{rebbapragada2009finding}. Automating the discovery of latent patterns in  real-world data can be challenging due to
noise. We focus on mitigating errata in pattern discovery from interval censoring~\cite{miller2011survival}. 

Interval censoring arises when time-series data are only observed within a known interval. 
The difference between an observed time value and the true time value, known as \textit{left-censorship}, and the lack of a common outcome against which to align, known as \textit{right-censorship}, can lead common techniques for clustering to erroneous conclusions about the underlying patterns.
To address this, practitioners must often manually align data
to a meaningful start time in order to find non-trivial groups via unsupervised clustering---a process whose difficulty can range from
expensive and time-consuming in some problems to infeasible in others. 
In this work, we develop a machine learning algorithm
that clusters time-series data while simultaneously correcting for interval censoring. In doing so, we automate the time-consuming
process of manual data alignment and use our method to reveal structure
that would otherwise not be found by straightforward application of clustering analysis. 

This paper is motivated by disease phenotyping in healthcare. Many diseases are biologically heterogeneous despite a common diagnosis---for example autism~\cite{Doshi-Velez2014}, heart failure~\cite{shah2015phenomapping}, 
and Parkinson's disease~\cite{fereshtehnejad2017clinical}. The variation in biomarkers (e.g., glucose or creatinine) across patients can stem from different patient subtypes that manifest in distinct disease trajectories. Scientists seek to understand this disease heterogeneity by identifying groups of people
whose biomarkers behave similarly.
For example, cardiologists use a measurement called ejection fraction as a heuristic to separate heart failure patients into two categories~\cite{owan2006trends}, 
with at least one of the two subtypes believed to be heterogeneous~\cite{shah2015phenomapping}. 
To better understand patient heterogeneity, cardiologists may turn to longitudinal, observational, and often irregularly-measured patient data for disease phenotype discovery.

\begin{figure*}[!t]
    \centering
    \subfloat[width=0.3\textwidth][]{
    \includegraphics[width=0.37\textwidth]{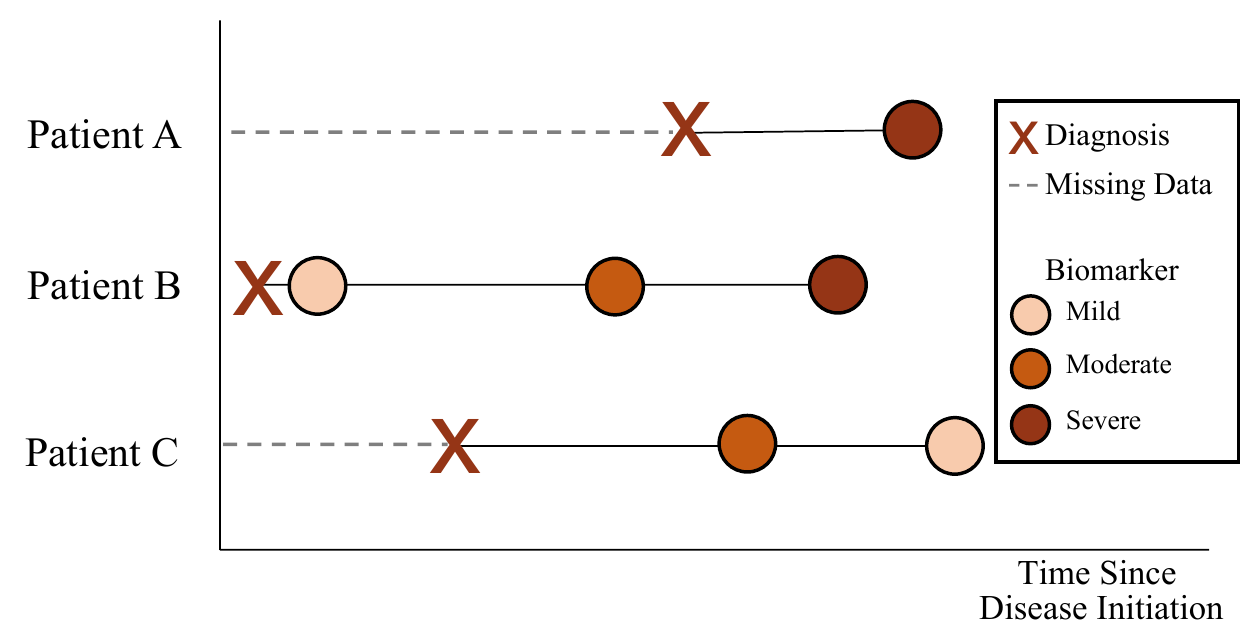}
    }
    \subfloat[width=0.4\textwidth][]{
    \includegraphics[width=0.3\textwidth]{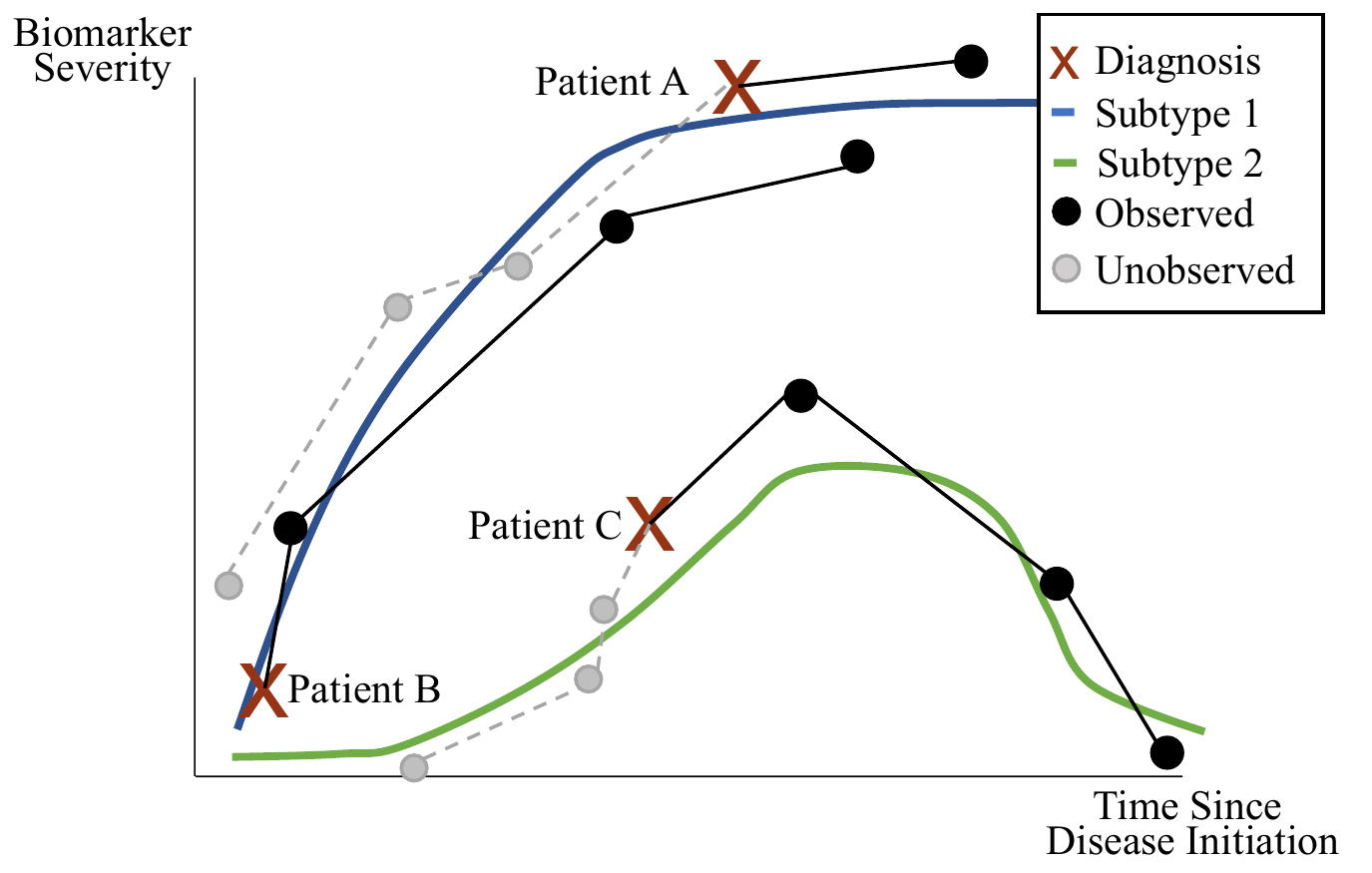}
    }
    \subfloat[width=0.1\textwidth][]{
         \includegraphics[width=0.2\textwidth]{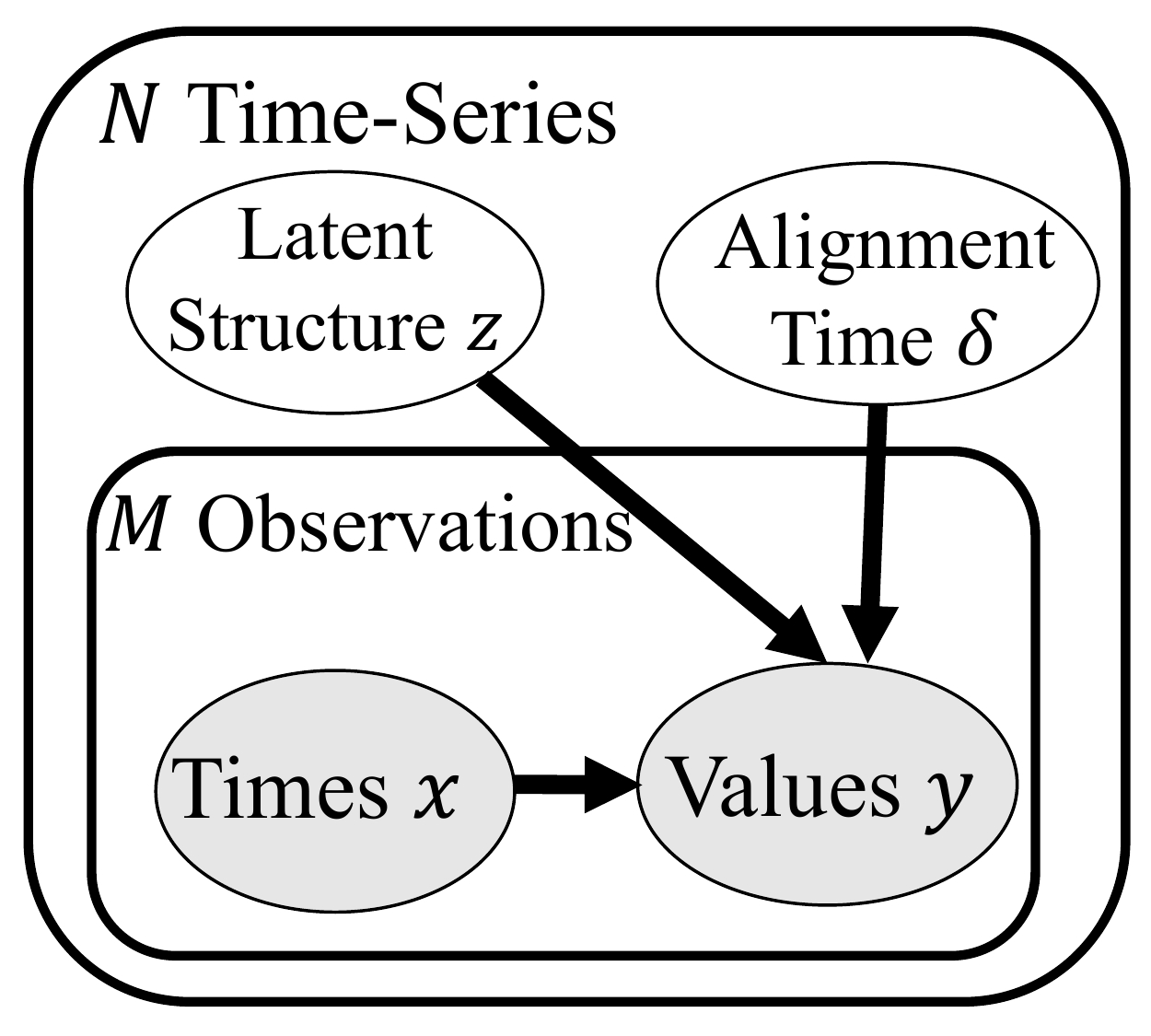}
         }
    \caption{
    \textbf{(a)} Patient data can be interval-censored, meaning longitudinal data can be missing based on entry to the dataset, e.g., first diagnosis, and the data may lack a common outcome against which to align. Patients may enter the dataset at any stage of the disease. \textbf{(b)} Interval censoring can make clustering patient time-series data challenging because data may be aligned incorrectly, e.g., first diagnosis. We seek to understand disease heterogeneity by inferring subtypes after correcting for misalignment. \textbf{(c)} Graphical model of SubLign.
    }
    \label{fig:illustration}
\end{figure*}

Interval censoring in healthcare data presents a significant challenge for disease phenotyping. Left-censorship in clinical datasets can occur when patients have \textit{delayed entry}, meaning data are unavailable before a diagnosis or first hospital visit. Factors including geographic proximity to a hospital~\cite{chan2006geographic}, financial access to care~\cite{miller2017health} or mistrust of the healthcare system~\cite{brandon2005legacy} can affect when a patient seeks medical help and consequently the beginning of data availability with respect to the underlying progression of their disease. 
Other datasets align patients by death time, but right-censorship restricts sample size to patients who have died. Since many factors affect mortality, other causes of death may confound results. 
For heart failure, a chronic disease that progresses over many years with a large range of onset ages and survival outcomes, this interval censoring can confound attempts to analyse disease heterogeneity using observational data.

For a simplified illustration of the problem, see Figure~\ref{fig:illustration}(a) which depicts the common reality of observational
health data whereas Figure~\ref{fig:illustration}(b) depicts the idealized latent substructure we would like to identify. 
Existing subtyping models applied to clinical data
assume (potentially erroneously) that patients are aligned at entry into the dataset or study.
The problem with such an assumption is that the disparity between true disease stage and observed observation time can result in unsupervised learning algorithms uncovering the wrong, or perhaps less interesting, structure. For example, a naive clustering algorithm might simply return clusters corresponding to the disease stage at entry into the study, which may simply recapitulate some of the biases mentioned earlier.


To that end, our paper makes the following contributions: 
\begin{enumerate}
    \item We introduce and formalize the problem of cluster recovery from interval-censored time-series data.
    \item We introduce a practical algorithm, SubLign, based on variational learning of a deep generative model, which:
        \begin{itemize}
        \itemsep0em
            \item Makes no assumption on the distribution of delayed entry alignment values besides extrema.
            \item Operates on multivariate time-series with varying lengths and missing values, both characteristics often found in real-world datasets.
            \item Is unsupervised, meaning that neither subtype labels nor alignment values are provided during training.
        \end{itemize}
    \item We prove an identifiability result showing that, in a noiseless setting, both the degree of delayed entry from left-censorship and the subtype identity are recoverable.
\end{enumerate}
We show robust quantitative results on synthetic data where, over multiple runs, our method outperforms baselines for subtyping and patient alignment. On two real-world clinical datasets---heart failure and Parkinson's disease---we automatically recover known clinical subtypes without manual alignment or expert knowledge.

\input{related}

\input{methodology}

\section{Evaluation Setup}
\label{sec:experiments}



\label{sec:synthetic_experiment_setup}
\paragraph{Synthetic data} We generate two classes of synthetic datasets from the \emph{sigmoid}
and \emph{quadratic} parameterizations in the `SubLign: Subtype \& Align' section. 
For the sigmoid dataset, we generate data from $K=2$ subtypes and $\forall i,m, |D_{i,m}|=3$ biomarker dimensions. For the quadratic dataset, we generate data with $K=2$ and $\forall i,m, |D_{i,m}|=1$.  See appendix for the data generation process for the six quadratic datasets. 


In both synthetic settings, we sample $N=1000$ patients with $M=4$ observations, variance $\sigma^2=0.25$, and max disease stage $T^+ = 10$.
For each patient, we sample subtype $s \sim \text{Bern}(0.5)$. The true disease stage is drawn $t_{m} \sim \text{Unif}(0,T^+)$ for observation $m \in \{1, \ldots, M\}$. The biomarker values are drawn $y_{m} \sim N(\lambda_{m}, \sigma^2)$ where $\lambda_{m} = \sum_{k \in \{1, \ldots,K\}} \mathbbm{1}(s_i = k) f_k(t_{m})$. 
For the sigmoid dataset, the first subtype generating function across three dimensions is $f_1(t) = [\sigma(-4+t), \sigma(-1+t), \sigma(-8+8t)]$ and the second subtype generating function is $f_2(t) = [\sigma(-1+t), \sigma(-8+8t), \sigma(-25 + 3.5t)]$. 
The observed disease time $x_{m}$ is shifted such that the first patient observation is at time 0. Therefore $x_{m} = t_{m} - \zeta$ where $\zeta = \min_{j \in \{1, \ldots, M\}}  t_{j}$ is the earliest true disease time for the patient.


\paragraph{Real-world clinical data} We study two real-world clinical datasets (Parkinson's disease and heart failure) to assess SubLign performance when data have realistic dynamics and are missing over time. Our clinical datasets inherently contain delayed entry due to external factors including patient access to healthcare. We include preprocessing code and internal review board details in the appendix. 

\textit{Heart failure (HF):} We use electronic health records from a large health system in the United States, currently redacted for double blind review. We identify patients who enter the emergency department with a diagnosis of HF and extract echocardiogram values---measurements from an ultrasound of the heart---from the full patient history. We include echocardiogram features that are present in more than 60\% of echo studies. Our dataset includes $N=1534$ patients and $|D_{i,m}|\leq 12;\forall i,m$ features with $M = 38$ maximum observations per patient over a potential span of 10 years in the dataset.
We extract 27 baseline features including race, sex, and comorbidities (e.g., renal failure) to validate subtypes only and not for use in the model.
The dataset values are linearly scaled such that values are between 0 and 1 with larger values denoting more abnormality. 

\textit{Parkinson's disease (PD): } We use publicly-available data from the Parkinson's Progression Markers Initiative (PPMI), an observational clinical study, totalling $N_t=423$ PD patients and $N_c=196$ healthy controls where $N=N_t+N_c$. We extract four biomarker measurements of autonomic, motor, non-motor, and cognitive ability from $N=619$ total participants with $M=17$ maximum observations per patient. Our baseline data include $25$ features including demographic information and patient history used to validate subtypes. For PD patients, the first recorded visit is within 2 years of the patient's PD diagnosis. Measurements are scaled between 0 and 1 with larger values corresponding to more abnormal values. We use the sigmoid parameterization of the SubLign model for both datasets because HF and PD are chronic and incurable diseases. 

\subsection{Hyperparameters and Baselines} 
\label{sec:experiment_setup}

We find optimal hyperparameters via grid search.
For both synthetic and clinical experiments, we search over hyperparameters including dimensions of the latent space $z$ (2, 5, 10), the number of hidden units in the RNN (50, 100, 200), the number of hidden units in the multi-layer perceptron (50, 100, 200), the learning rate (0.001, 0.01, 0.1, 1.), regularization parameter (0., 0.1, 1.), and regularization type (L1, L2).
We select the hyperparameter configuration with the best validation loss, as measured by Equation~\ref{eqn:var_lwbd}. 
Our models are implemented in Python 3.7 using PyTorch ~\cite{paszke2019pytorch} and are learned via Adam~\cite{kingma2014adam} on a single NVIDIA k80 GPU for 1000 epochs. 
We set alignment extrema $\delta^+=10$ based on the maximum of the synthetic dataset and the maxima of the HF and PD datasets. 
We search over 50 time steps with $\epsilon=0.1$.

We compare to seven different baselines. 
Our greedy baseline, denoted as KMeans+Loss, first clusters the observed values using $k$-means clustering. Then, using the inferred labels $s$, we simultaneously learn $\theta_k$ for each subtype and $\delta_i$ for each patient by minimizing:
\begin{equation}
    \arg \min_{\theta, \delta} \sum_{i=1}^N \sum_{j=1}^M \sum_{d=1}^D\sum_{k=1}^K \mathbbm{1}[s_i = k] [y_{i,j,d} - f(x_{i,j,d} + \delta_i; \theta_k)]^2
\end{equation}
using Broyden–Fletcher–Goldfarb–Shanno. This naive clustering based approach (in the space of the original data) attempts to correct for shifts in time.
We also compare to:
\begin{enumerate}
\itemsep0em
    \item SubNoLign: a modified SubLign with no alignment value. This model is comparable to \cite{zhang2019data}, which learns a deep patient representation while controlling for model architecture
    \item SuStaIn~\cite{young2018uncovering}: a subtype and stage inference algorithm for disease progression,
    \item  BayLong: a Bayesian model of longitudinal clinical data \cite{huopaniemi2014disease},
    \item PAGA~\cite{wolf2019paga}: a state-of-the-art single-cell trajectory pseudo-time method,
    \item Clustering with dynamic time warping (DTW) using kernel methods~\cite{cuturi2011fast} and soft-DTW~\cite{cuturi2017soft},
    \item SPARTan: A tensor factorization based approach for phenotyping from time-series data~\cite{perros2017spartan}
\end{enumerate}
We detail baseline implementations in the appendix.

\subsection{Evaluation} 
\label{sec:evaluation}
We evaluate models on 5 trials, each with a different randomized data split and random seed. For each trial, we learn on a training set (60\%), find the best performance across all hyperparameters on the validation set (20\%), and report the performance metrics on the held-out test set (20\%). The same data folds are used across all models for each trial.

We report the performance on the test set over three metrics.
\textit{Adjusted Rand index (ARI)} measures whether pairs of samples are correctly assigned in the same or different subtypes ~\cite{hubert1985comparing}.
The \textit{Swaps metric} reports the number of swaps needed to sort the predicted disease times into the true disease stages, expressed as percentage of total possible swaps, with full equation in the appendix.
The \textit{Pearson correlation coefficient} expresses correlation between the predicted and true disease stage.
ARI measures the clustering performance while the Swaps metric and the Pearson correlation coefficient quantify how well the learning algorithm infers the alignment values. 

\begin{figure*}
    \centering
\label{tab:sigmoid_results}

\subfloat[width=0.8\textwidth][]{
\begin{small}
\begin{tabular}{lrrr}
\toprule
{\sc Model} & {\sc ARI}	$\uparrow$ & {\sc Swaps}	$\downarrow$ & {\sc Pearson}	$\uparrow$ \\
\midrule
SubLign & \textbf{0.94 $\pm$ 0.02} & \textbf{0.09 $\pm$ 0.00} & \textbf{0.85 $\pm$ 0.04}\\
SubNoLign & 0.81 $\pm$ 0.21 & --  & -- \\
KMeans+Loss &  0.67 $\pm$ 0.04 & 0.21 $\pm$ 0.03 & 0.49 $\pm$ 0.01\\
SuStaIn &   0.66 $\pm$ 0.02 & 0.16 $\pm$ 0.00 & 0.30 $\pm$ 0.02\\
BayLong &  0.19 $\pm$ 0.18 & 0.48 $\pm$ 0.00 & 0.01 $\pm$ 0.02\\
PAGA &  0.32 $\pm$ 0.05 & 0.52 $\pm$ 0.07 & 0.04 $\pm$ 0.20\\
Soft-DTW & 0.06 $\pm$ 0.01 & -- & -- \\
Kernel-DTW & 0.06 $\pm$ 0.07& -- & --\\
SPARTan & 0.22 $\pm$ 0.18 & -- & -- \\
\bottomrule
    \end{tabular}
\end{small}
}
\subfloat[][]{
\begin{small}
\begin{tabular}{l r r r}
\toprule
{\sc Model} & {\sc ARI}\\
\midrule
\textbf{SubLign} & \textbf{0.58 $\pm$ 0.12} \\
SubNoLign & 0.42 $\pm$ 0.14 \\
KMeans+Loss & 0.05 $\pm$ 0.04 \\
SuStaIn & 0.12 $\pm$ 0.11 \\
BayLong & 0.04 $\pm$ 0.17 \\
PAGA & 0.02 $\pm$ 0.02 \\
Soft-DTW & 0.46 $\pm$ 0.43 \\
Kernel-DTW & 0.21 $\pm$ 0.36 \\
SPARTan & 0.15 $\pm$ 0.10 \\
\bottomrule
\end{tabular}
\end{small}
}
\caption{Means and standard deviations over 5 trials for: \textbf{(a)} synthetic sigmoid dataset with 1000 patients, 3 dimensions, and 4 observations per patient, \textbf{(b)} 619 patients in the PPMI dataset including 423 Parkinson's disease patients and 196 healthy controls. Baseline methods include SuStaIn~\cite{young2018uncovering}, BayLong~\cite{huopaniemi2014disease}, PAGA~\cite{wolf2019paga},
Soft-DTW~\cite{cuturi2011fast},
Kernel-DTW~\cite{dhillon2004kernel}, and SPARTan~\cite{perros2017spartan}.
}
\label{fig:results}
\end{figure*}

\paragraph{Quantitative metrics on clinical datasets:} Because real world data often lack ground truth labels for subtype or alignment, we create two semi-synthetic experiments with clinical datasets. For HF, we evaluate SubLign's ability to infer relative disease stage by introducing additional censoring into the test sets. Specifically, we train SubLign using 80\% data (train and validation data) as usual. We then modify the remaining data (20\%) by removing the first year of patient observations, creating distorted test set $(X',Y')$, and by removing the last year of patient observations, creating $(X'',Y'')$. The same amounts of observations are removed from each set to control for length of observations.
We infer alignment values using the trained SubLign model: $\delta'$ from $(X',Y')$ and 
$\delta''$ from $(X'',Y'')$. By construction, $\delta' > \delta''$. We report the percentage of patients for which 
SubLign is able to recover this relationship.
For PD, we report the held-out clustering performance for healthy control patients and patients with PD. We use disease status (PD patient or healthy control) as labels and $K=2$ subtypes.






\section{Evaluation and Analysis}


\subsection{Recovering Subtypes with interval censoring}
\label{sec:synthetic_results}

SubLign is able to recover subtypes despite interval censoring, outperforming all baselines. For sigmoid synthetic data (Figure \ref{fig:results}(a), ARI column), SubLign can recover subtypes (mean ARI of 0.94) better than the KMeans+Loss baseline (0.67) which assumes a greedy approach. 
Not correcting for alignment time decreases the quality
of inferred subtypes as can be seen in SubNoLign (0.81). 

Similarly, SubLign recovers known subtypes in the PD dataset with statistically significantly higher ARI over baselines (see Figure~\ref{fig:results}(b)). When ARI performance intervals overlap, we use a t-test on pairwise differences over trials to compute statistical significance. We specify the optimal parameters for experiments and further describe statistical significance methods in the appendix.

Some baselines appear to suffer because SubLign leverages the longitudinal nature of patient data compared to the cross-sectional assumptions of PAGA and SuStaIn. Other baselines have strong priors, i.e., \cite{huopaniemi2014disease}, which may explain its poor performance. Dynamic time warping methods appear to perform poorly for datasets with few observations or high missingness rates. Lastly, the tensor factorization method~\cite{perros2017spartan} fails, likely because transforming data from continuous to discrete time results in a very large and extremely sparse matrix factorization that is a tricky optimization problem. 

We include additional results in the appendix, including visualizations of the SubLign subtypes compared to baselines, model misspecification analysis, and experiments varying the level of missingness.

\subsection{Recovering Known Alignment Values}
\label{sec:known_alignment}

For the synthetic sigmoid data, SubLign outperforms baselines in inferring alignment values (Figure~\ref{fig:results}(a), Swaps and Pearson columns). SubLign recovers alignment values better according to the Swaps metric (mean value of 0.09) and the Pearson metric (mean value of 0.85) compared to the next best baselines of KMeans+Loss and SuStaIn. Note that many baselines only recover subtypes and do not learn patient alignment values.
Although the real-world clinical datasets do not contain true alignment values, we use the previously described HF setup with an artificially censored test set. We find that SubLign predicts known alignment relationships in an altered dataset. When evaluated on the manipulated test data, SubLign recovers the constructed relationship of $\delta' > \delta''$ with a higher performance ($71\% \pm 2\%$) over five trials compared to K-Means+Loss ($57.8 \pm 4\%$) and SuStaIn ($53.8 \pm 3\%$). See appendix for full results.


\subsection{Clinical Insights from Correcting for Misalignment} 
\label{sec:clinical_insights}

We validate SubLign subtypes learned from the HF and PD datasets using known clinical findings. 
The appendix includes a table of baseline features, which are not included as input to SubLign, with statistically significant differences in subtypes: 7 features (out of 26) for PD and 11 features (out of 27) for HF. 

\paragraph{Heart failure} Cardiologists classify patients into two groups based on ejection fraction: HF with reduced ejection fraction (systolic HF) and HF with preserved ejection fraction (diastolic HF). However, in our HF dataset, over 30\% of patients correspond to neither group based on clinical diagnosis. For SubLign, we set $K=3$, one more than the number of known groups, to discover a potentially new subtype. Without ground truth subtype labels, we observe that SubLign finds systolic HF and diastolic HF as statistically significant baseline features. Of the three subtypes, subtype C corresponds to systolic HF, and subtype A and B correspond to diastolic HF, mirroring known clinical heterogeneity in diastolic HF~\cite{shah2015phenomapping}. Of the two diastolic HF subtypes, subtype A has a higher proportion of women while subtype B has a higher rate of obese patients, both subgroups with documented heterogeneity in diastolic HF~\cite{duca2018gender,tadic2019obesity}. In contrast, the subtypes found by the KMeans+Loss baseline do not include known systolic HF and diastolic HF as statistically significant features. See appendix for full results.


\paragraph{Parkinson's disease} Biomarkers used to track PD are self-reported, which can be biased, subjective, and noisy. Despite these obstacles, SubLign discovers subtypes that match known clinical findings on two cohorts: 619 combined PD and healthy control patients, and 423 PD patients only.  

For PD and healthy control patients, we run SubLign with $K=2$ to uncover characteristics of the two known groups. SubLign subtype A clearly corresponds to healthy controls whereas subtype B designates PD patients. Statistically significant baseline features include all components of the University of Pennsylvania Smell Identification Test (UPSIT), which is a measure of smell dysfunction and highly linked to PD~\cite{haehner2009prevalence}, and having a full sibling or biological dad with PD, which aligns with research suggesting PD may be hereditary~\cite{klein2012genetics}.  
For PD patients only, we set $K=3$ in SubLign to explore and discover potential disease heterogeneity. Statistically significant baseline features include race and gender, which parallel recent clinical findings about heterogeneity in PD manifestation~\cite{taylor2007heterogeneity,duca2018gender} and indicate new potential areas for future work. See appendix for tables of statistically significant baseline features stratified by the discovered subtypes for HF and PD. 

\section{Discussion}



We study the task of clustering interval-censored time-series data. We present our method, SubLign, to learn latent representations of disease progression that correct for temporal misalignment in real-world observations and consider conditions for identifiability of subtype and alignment values. Empirically, our method outperforms seven baselines, and analysis of subtypes reveals clinically plausible findings. Better modeling of disease heterogeneity through alignment can help clinicians and scientists to better understand and predict how chronic diseases with many subtypes may progress. We hope that our model---in learning a continuous latent space to model heterogeneity---may be applied to other domains where subtypes and temporal alignment are entangled, for example gene expression analysis~\cite{bar2002new} or cancer pathways~\cite{wu2015inference}.

Our model introduces directions for future work. 
Practically, SubLign assumes that $\delta$ and $z$ are marginally independent. Intuitively, this means that, across all subtypes of a disease, the time at which the patient enters the data
is independent of any other factor.
There are certainly cases where this assumption may easily be violated, and therefore it remains an important area for further improvement. 
Our results for theoretical and noiseless identification do not naturally extend to the noisy setting since root finding, without further assumptions, can be sensitive to noise in the coefficients of the polynomials. Alternative strategies for identification that generalize our result are fertile ground for future work.



\section*{Acknowledgements}
The authors thank Steven Horng for assistance with the heart failure dataset and Christina Ji for assistance with the Parkinson's disease dataset. The authors thank Steven Horng and Rahul Deo for clinical guidance on heart failure experiments. The authors thank Rebecca Boiarsky, Hunter Lang, and Monica Agrawal for helpful comments. Data used in the preparation of this article were obtained from the Parkinson’s Progression
Markers Initiative (PPMI) database. PPMI --- a public-private partnership --- is funded by The Michael J. Fox Foundation for
Parkinson’s Research and funding partners. This work was supported by NSF CAREER award \#1350965.

\bibliography{references}
\bibliographystyle{apalike}

\appendix
\input{supplement}
\end{document}

%% file: related.tex
\section{Related Work}

Learning alignment and clustering has been studied in fields across computer vision, signal processing, and health. Approaches often make assumptions including few discrete time steps~\cite{young2018uncovering,huopaniemi2014disease}; a single piecewise linear function~\cite{young2018uncovering} or Gaussian mixture model~\cite{huopaniemi2014disease}; 
significantly more samples per object than number of objects~\cite{mattar2012unsupervised, gaffney2005joint}; 
very small windows of potential misalignment~\cite{liu2009simultaneous, listgarten2007bayesian}; or known lag time~\cite{li2011dynamic}. Methods that directly measure similarity between time-series, e.g., dynamic time warping~\cite{cuturi2011fast} or methods that aggregate multiple imputation methods~\cite{faucheux2021clustering} can also be used for clustering time-series data.
Our method aims to cluster interval-censored multivariate time series without these constraints.

Clinicians and scientists learn disease subtypes to better understand heterogeneity in disease progression in a process known as disease phenotyping. Existing approaches often rely on the assumption that the observed measurements are aligned  --- and therefore not censored. Researchers then apply clustering techniques like hierarchical clustering of time series~\cite{Doshi-Velez2014}, affinity clustering~\cite{luo2020multidimensional}, or matrix factorization~\cite{udler2018type,perros2017spartan}. Other models define disease subtypes as stages of disease progression~\cite{alaa2019attentive}. For this work, we define disease subtypes as distinct from disease stage and jointly learn both.

%% file: methodology.tex
\section{SubLign: Subtype \& Align}
\label{sec:model}

There are two stages to SubLign. First, we learn a generative model of the observed data which disentangles variation in the observed data due to delayed entry from variation related to subtype identity. Second, we infer subtype representations and (optionally) cluster the representations to obtain the explicit subtype identity for each time-series. Figure~\ref{fig:illustration}(c) describes the graphical model, and Algorithm 1 depicts the pseudocode for this procedure.

\begin{algorithm}[t]
   \caption{SubLign}
   \label{alg:sublign}
\begin{algorithmic}[1]   \STATE {\bfseries Input:} Observation times $X \in \mathbb{R}^{N \times M}$, biomarkers $Y \in \mathbb{R}^{N \times M \times D}$
      \STATE{\bfseries Output:} $\tau_k$ for each subtype and  $\hat{\delta_i}$ for each patient
   \STATE {\bfseries \underline{Step $1$: Learning}} 
  \REPEAT
   \STATE Encode time-series: $h_i = \text{RNN}([X_i, Y_i]) \; \forall i \in \{1, \ldots, N\} $
   \STATE Compute variational distribution $q(Z_i|X_i,Y_i)= \mathcal{N}(\mu(h_i;\phi_2), \Sigma(h_i;\phi_3)) \forall i \in \{1, \ldots, N\} $ 
   \FOR{patient $i=1$ {\bfseries to} $N$}
   \STATE Run grid-search to find $\hat{\delta_i} = \arg\max_{q(\delta_i)} \mathcal{L}(Y_i | X_i;\gamma,\phi,q(\delta_i))$ (Eq.~\ref{eqn:var_lwbd})
   \ENDFOR
   \STATE Update $\gamma,\phi$ via stochastic gradient ascent on $\mathcal{L}(Y|X;\gamma,\phi, \hat{\delta})$ 
  \UNTIL{convergence}
    \STATE {\bfseries \underline{Step $2$: Inference and Clustering}}
    \STATE Infer $\mathcal{Z} = \{z_i| z_i = \mu(h_i; \phi_2) \}$ for $X_i,Y_i$
   \STATE Find $K$ clusters using $k$-means on $\mathcal{Z}$ and compute cluster centers $\mu_k$
   \STATE Infer parameters of subtype trajectories $\tau_k = g(\mu_k)$
\end{algorithmic}
\end{algorithm}




\paragraph{SubLign Generative Model} 

Consider the following setup. We observe $N$ multivariate time-series (one for each patient), each of length up to $M$: $[(x_{1,1},y_{1,1}), \ldots, (x_{1,M},y_{1,M})], \ldots [(x_{N,1},y_{N,1}), \ldots, \\(x_{N,M},y_{N,M})]$. $y_{i,m}\in\mathbb{R}^D$ is a vector of observations for time-series $i$ at time-stamp $x_{i,m}\in\mathbb{R}^+$. We denote collections of observations as $Y_i = \{y_{i,1},\ldots,y_{i,M}\}$ and time-stamps as $X_i= \{x_{i,1},\ldots,x_{i,M}\}$ for patient $i$.

Figure~\ref{fig:illustration}(c) depicts the graphical model corresponding to the latent-variable generative model of continuous-time multivariate data:

\begin{align}
\label{eqn:gen_process}
    &\forall i=\{1,\ldots,N\},\;\forall m\in\{1,\ldots,M\}, \nonumber \\
    &\forall d\in D_{i,m}, \; \delta_i\sim \text{Cat}(\mathcal{D}),
    z_i \sim \mathcal{N}(\mathbf{0},\mathbb{I}),\; 
    \Theta = g(z_i;\gamma), \nonumber\\ &\overline{y}_{i,m}[d] = f(\kappa(x_{i,m} + \delta_i; \Theta[d])),\;
    y_{i,m} \sim \mathcal{N}(\overline{y}_{i,m}[d],  \mathbf{I})
\end{align}

We drop indices denoting patient and dimension where unnecessary. $D_{i,m}$ denotes the set (and $|D_{i,m}|$ denotes the number) of observed
biomarkers for patient $i$ at their $m$-th observation. 
To accommodate missing data, not all biomarkers are required to be measured at every observation.
Each delayed entry value $\delta_i \in \mathbb{R^+}$ has a maximum alignment deviation value $\delta^+$ over all time-series. We discretize the closed interval $[0,\delta^+]$ as $\mathcal{D} = [0, \epsilon, 2\epsilon,\ldots, \delta^+]$ with
hyperparameter $\epsilon$ and use a categorical distribution over $\mathcal{D}$ with uniform probabilities over each element as our prior over $\delta_i$. Function $g:\mathbb{R}^{N_z}\to\mathbb{R}^{D\times (P+1)}$ 
has parameters $\gamma$ and maps from the latent variable to $\Theta\in\mathbb{R}^{D\times (P+1)}$, a matrix of parameters 
for $D$ polynomials, each of degree $P$. 
$f$ is a known link function that describes how observed values relate to observed time-points.

\textbf{Parameterization} 
We discuss the specific parameterizations of Equation \ref{eqn:gen_process}. In the context of our motivating application of disease phenotyping, these functions represent common characteristics in the progression of patient biomarkers.

\textit{Link function and polynomials:} For $f$ and $P$, we study the following choices: \textit{Sigmoid}: $P=1$ and $f(x)=\frac{1}{1+\exp(-x)}$ and \textit{Quadratic}: $P=2$ and $f(x)=x$.
The sigmoid function can represent bounded and monotonically increasing clinical variables. 
The quadratic function represents cases where disease severity, as measured by biomarkers, decreases (likely in response to therapy) and then increases (once therapy fails), or vice versa.
Other choices for $P, f$ are permissible as long as they are differentiable with respect to the model parameters. We allow for the possibility that individual biomarkers have different parameterizations.
 
\textit{Modeling polynomial parameters:} We parameterize $g(z_i;\gamma)$ using a two layer neural network with ReLU activation functions with parameters $\gamma$. 
To be concrete, if $D=1, P=2$, and $f$ is the sigmoid function, then the outputs of $g$ are $[\beta_0(z), \beta_1(z)]$ and $y=\frac{1}{1+\exp^{-(\beta_0(z)x + \beta_1(z))}}$. Similarly, if $D=1,P=1$, and $f$ is the quadratic function
then the outputs of $g$ are $[a(z),b(z),c(z)]$ and $y=a(z)x^2+b(z)x+c(z)$.

\subsection{Step $1$: Learning}
We learn the parameters $\gamma$ of the model in Equation \ref{eqn:gen_process} via maximum likelihood estimation. Since the model is a non-linear latent variable model, we maximize a variational lower bound on the conditional likelihood of data given the time-stamps corresponding to observations.
\begin{align}
&\log \prod_{m=1}^M \prod_{d \in D_{i,m}} p(y_{i,m}[d] | x_{i,m}; \gamma) = \log p(Y_i|X_i;\gamma) \\
&\geq \mathcal{L}(Y_i|X_i;\gamma,\phi)\nonumber\\ &=\mathbb{E}_{q(Z_i|X_i,Y_i;\phi)}\Big[ \log \sum_{\delta_i} p(Y_i|X_i,\delta_i,Z_i;\gamma) p(\delta_i) + \nonumber\\
&\quad \quad \quad +\log \frac{p(Z_i)}{q(Z_i|X_i,Y_i;\phi)}\Big]
\geq \mathcal{L}(Y_i|X_i;\gamma,\phi,q(\delta_i))\nonumber\\ 
&= \mathbb{E}_{q(Z_i|X_i,Y_i;\phi)} \Big[  \mathbb{E}_{q(\delta_i)} \Big [\log p(Y_i|X_i,\delta_i,Z_i;\gamma) \nonumber\\ 
&\quad \quad \quad + \log \frac{p(\delta_i)}{q(\delta_i)} \Big ]   + \log \frac{p(Z_i)}{q(Z_i|X_i,Y_i;\phi)}\Big]\label{eqn:var_lwbd}
\end{align}
The first lower bound uses a variational distribution for $Z$ parameterized via an inference network \cite{kingma2013auto,rezende2014stochastic} with parameters $\phi$.
The second lower bound is a variational distribution over $\delta$. The function $q(\delta_i)$ parameterizes the space of one-hot distributions in $\mathcal{D}$, i.e., a categorical 
distribution over discrete choices of $\delta_i$. 

Our learning algorithm alternates between two steps.
We first maximize the lower bound using subgradient ascent. To do so, we solve: 
$\hat{\delta_i} = \arg\max_{q(\delta_i)} \mathcal{L}(Y_i | X_i;\gamma,\phi,q(\delta_i))$. For our choice of variational distribution, this maximization can be performed via a grid search.
We then 
derive gradients $\nabla_{\gamma,\phi} \mathcal{L}(Y_i | X_i;\gamma,\phi,\hat{\delta}_i)$ 
to update the generative model and inference network via stochastic gradient ascent.

\subsection{Step $2$: Obtaining Learned Subtypes} 
After learning the model, we may re-use the inference network
to predict the latent variable $z_i = \mu(h_i;\phi_2)$ for each patient in the training set. Combining $z_i$ across all time-series 
gives us the set $\mathcal{Z}$. When reasonable, we refer to $\mu(h_i; \phi_2)$ as $\mu_i$.


Although latent variable $z_i$ encodes latent structure from each time-series, we may be interested in explicit subtypes for a given value of $K$. To obtain discrete subtypes, we can run clustering algorithms on $\mathcal{Z}$ to obtain $K$ cluster centers $\{\mu_1,\ldots, \mu_K\}$. Because we use a Gaussian prior for our biomarker values, measuring distances in the space of the latent variable can be done with the Euclidean norm, making the $k$-means a reasonable choice of clustering algorithm. 

We compute $\{\tau_1,\ldots,\tau_K\}$ where $\tau_k = g(\mu_k)$
as the progression-patterns corresponding to each of the discrete subtypes of the disease. For example, if $f\circ \kappa$ is linear, then we obtain $K$ different
slopes and biases, each of which describes how the time-series
behaves in that subtype. In practice, $K$ may be chosen based on domain knowledge; alternatively, qualitative results can be assessed for each version of $K$, e.g., by plotting the corresponding $f$ functions.

\subsection{Remarks on SubLign}

\textit{Role of the latent variable:} The latent variable $z$ plays an important role in quantifying how each biomarker behaves.
Each time-series's latent variable is used to predict the parameters of $D$ polynomial functions and $f\circ\kappa$ maps from observation times onto the observed biomarkers.
Time-series whose representation space $z$ are \emph{close} hail from the same subtype, and consequently manifest similar patterns in their biomarkers. This variation in $z$ results in variation in the parameters $\Theta$ and therefore in variation among the data as a function of the time-points. 
As an illustration, in Figure \ref{fig:illustration}(b) in the blue phenotype, $f$ is the sigmoid function. Depending on the latent space, we could imagine a one-dimensional $z$ where $z<0$ represents the curve (and subtype) in blue and $z\geq0$ to represent the curve in red. 
The value of $\delta_i$ indicates the degree of delayed entry associated with each time-series.
The delayed entry from interval censoring is corrected by applying the scalar $\delta_i$ element-wise to $X_i$ and then transforming it by $f$. 

\textit{Scalability:} The runtime of SubLign is impacted by the grid search over model parameter $\delta_i$. The corresponding lines 7-9 in Algorithm 1 have complexity $\mathcal{O}(NSF)$ where $F$ is the complexity associated with a single forward pass of the inference network and the generative model, $N$ is the number of examples, and $S =\delta^+ / \epsilon$  is the number of time steps. The model practitioners may therefore balance computational resources with $S$. In our experiments, we found comparable performance for $S$ as low as 5. 

\textit{Real-world clinical data:} SubLign is motivated by, and designed to capture, variation in clinical biomarkers while taking into account the challenges of clinical data. Observational healthcare data are often irregularly spaced, and contains missingness. The use of a continuous time model allows us to naturally handle the former issue since we only maximize the likelihood of data corresponding to time-points where they are observed. When a single biomarker is missing while others are observed, it may be marginalized out (by ignoring the corresponding loss term). 
r
\textit{Accommodating different kinds of censorship:} Equation \ref{eqn:gen_process} naturally characterizes delayed entry arising from left-censorship. SubLign can also accommodate right-censorship by reversing the sequence of time-series and applying Algorithm 1 (resulting in our ability to infer the degree of right censorship). When both left-censorship and right-censorship are present, it corrects for left-censorship explicitly (using $\delta$) while right-censorship is implicitly accounted for since we only maximize the likelihood of data up to the point that we observe time-series.

\input{identifiability}

%% file: identifiability.tex
\section{Identifiability under a noiseless model}
\label{sec:identifiability}

While SubLign presents a viable, practical model for clustering and aligning censored time-series data, it is worth reflecting upon whether we can ever identify subtype and alignment from observational data. In what follows, we present theoretical conditions that show that there exists conditions under which the problem we study is identifiable. 


\paragraph{Identifiability} We assume distinct time stamps for the $M$ observations in $X_i$. The generative process we assume for $Y_i$, conditional on $X_i$, is: 
\begin{align}
    \label{eqn:identifiable_model}
    &\forall i=\{1,\ldots,N\}, \; s_i \sim \text{Cat}(K),\;  
    \nonumber \\
    &\forall m\in\{1,\ldots,M\},\;d \in D_{i,m}, \nonumber \\
    &y_{i,m}[d] = f(\kappa(x_{i,m} + \delta_i; \theta^P[s_i,d])) 
\end{align}
where $s_i \in \{1,\ldots, K\}$ is the subtype for time-series $i$, $D_{i,m}$ denotes the set of all observations at time-step $m$ for time-series $i$ where $\forall i,m, |D_{i,m}| \leq D$, and $\delta_i$ is the delayed entry value.
The link function $f: \mathbb{R} \to \mathbb{R}$ has no parameters whereas $\kappa: \mathbb{R} \to \mathbb{R}$ is an unknown polynomial function of degree $P \in \mathbb{Z}^+$. We denote the parameters of $\kappa$, for each subtype and dimension (e.g., biomarker), as $\theta^P \in \mathbb{R}^{K \times D \times (P+1)}$. 
We denote $\theta^P[s_i,d]$ as selecting the $(s_i,d)$-th vector of size $P+1$ from the tensor $\theta^P$. Similarly, $\theta^P[s_i]$ selects the $s_i$-th matrix of size $X \times (P+1)$. We define $\theta_p$ as the $p$-th coefficient of any polynomial function parameter set $\theta$. 

By construction, we have that
$s_i = s_{i'}\iff \theta^P[s_i] = \theta^{P}[s_{i'}]$, i.e., for two subtypes, the values of each time-series are described either by $y = f(\kappa(x;\theta^P[s_i]))$ or $y = f(\kappa(x;\theta^{P}[s_{i'}]))$.  


We begin with a set of assumptions for identifiability. 


\begin{assumption}
$f$ is invertible, and $\kappa(x,\theta) = \theta_{0} +\sum_{p=1}^P \theta_p x^{p}$ describes a family of polynomial functions in x with parameter $\theta$ and degree $P>0$. The parameters of each subtype are unique.
\label{assum:functionals}
\end{assumption}


\begin{assumption}
$M\geq P+1$, \, i.e., for each patient time-series, there exists at least one of the $D$ features where we observe at least $P+1$ values. 
\label{assum:enough_obs}
\end{assumption}

\begin{assumption}
For each subtype $s_k$, there exists a time-series $i$ whose alignment $\delta_i=0$, meaning no delayed entry.
\label{assum:delta0}
\end{assumption}

\begin{theorem}
\label{thm:identifiability}
Under assumptions \ref{assum:functionals}, \ref{assum:enough_obs}, \ref{assum:delta0} for the model in Equation \ref{eqn:identifiable_model}, we can identify the time-delays $\delta_1,\ldots, \delta_N$. We can identify
the polynomial coefficients $\theta^P$ up-to a permutation of its rows and columns and the identity of $s_1,\ldots,s_N$ up-to a permutation over $K$ choices. 
\end{theorem}

\textbf{Proof sketch: } We defer the full proof for Theorem \ref{thm:identifiability} to Appendix A but provide the sketch here. Consider the case where we have a single biomarker for each patient. The proof is constructive; first we transform the data using the inverse of $f$ resulting in a set of data drawn from polynomial equations. The polynomial coefficients may be estimated from the observed data; we can then find the roots of these polynomials and pick the smallest root. These roots exactly quantify the degree of delayed entry; i.e. they tell us how much each polynomial has been shifted by. We can correct each time-series for this shift, re-estimate polynomial coefficients from the shifted-polynomials and cluster them to reveal the underlying subtype identity for each time-series (and consequently each patient). 


\textbf{Remarks:} Theorem \ref{thm:identifiability} describes conditions under which
delayed entry and the polynomial parameters of cluster biomarker progression are identifiable. This encouraging result demonstrates scenarios where the parameters of the model in Equation \ref{eqn:identifiable_model} can provably be identified.


\textit{On the assumptions for identification:} It is possible to relax Assumption \ref{assum:delta0} to only require the existence of a single time-series from each subtype; this modification only allows identifiability of $\delta_1,\ldots,\delta_N$ up to a translation within each subtype. The above result relies on the existence of at least one biomarker for which there are sufficiently many
observations -- this is a reasonable assumption in the context of clinical data since there
is often a \emph{canonical} biomarker tracked over time for each disease.

\textit{On the strategy for identification:} We conjecture our analysis for identification of subtype and alignment is of independent interest for identifying causal effects in survival analysis where an important challenge is how to handle confounding that jointly affects both survival time and censorship. Related work in this field \cite{seaman2020adjusting,choi2017estimating} has focused on restricting the class of models used to characterize the survival function. Our work presents distinct parameteric assumptions towards this goal.

%% file: supplement.tex
\clearpage
\appendix
\section*{Appendix}

\begin{itemize}
\itemsep0em
    \item In Section~\ref{sec:app_identifiability}, we present the proof for conditions of identifiability and variational lower bound derivation.
    \item In Section~\ref{sec:app_experiment_setup_details}, we describe additional experiment setup details for the sigmoid, quadratic, and clinical experiments.
    \item In Section~\ref{sec:app_add_experiment_results}, we present additional experimental results on discovering subtypes including visualization of subtypes, model misspecification, and missingness methods. 
    \begin{itemize}
    \itemsep0em
        \item Visualizations of subtypes show that SubLign subtypes match the data generating function closer than SubNoLign. 
        \item Model misspecification results show that SubLign is robust to misspecification from cubic spline data generation. 
        \item Missingness method experiments show that SubLign outperforms baselines for various missingness rates with comparable performance to some baselines for no missingness.
    \end{itemize}
    \item In Section~\ref{sec:quadratic_data_results}, we present quadratic experiment results.
    \begin{itemize}
    \itemsep0em
        \item An additional six synthetic experiments show that SubLign outperforms baselines.
        \item In cases where alignment values are not identifiable (e.g., one subtype's data generating function is a flat line), SubLign cannot recover those alignment values.
    \end{itemize}
    \item In Section~\ref{sec:data_privacy_ethics}, we discuss internal review board and patient privacy details for the clinical datasets used.
    \item In Section~\ref{sec:app_clinical_insights}, we present the set of statistically significant features for the heart failure and Parkinson's disease clinical experiments.
        \begin{itemize}
        \itemsep0em
        \item Model misspecification results show that SubLign is robust to misspecification from cubic spline data generation. 
        \item Missingness method experiments show that SubLign outperforms baselines for various missingness rates with comparable performance to some baselines for no missingness.
    \end{itemize}
    
\end{itemize}

\section{Identifiability and Inference}
\label{sec:app_identifiability}
\subsection{Identifiability}

\input{identifiability_proof_2}

\subsection{Variational Lower Bound}

\begin{align}
\label{eqn:lw_bd}
    \log &\; p(Y|X;\gamma)\nonumber \\
    &=\log  \int_{Z,\delta} p(Y,Z,\delta|X;\gamma) \, dZ \, d\delta \nonumber \\ 
    &=\log \int_{Z,\delta}q(Z|X,Y;\phi) \nonumber  \frac{p(Y,Z,\delta|X;\gamma)}{q(Z|X,Y;\phi)} \, dZ \, d\delta \nonumber\\
    &\geq  \int_{Z,\delta} q(Z|X,Y;\phi) \log\frac{p(Y,Z,\delta|X;\gamma)}{q(Z|X,Y;\phi)}\, dZ \, d\delta  \\
    &= \int_{\delta} q(\delta) \int_{Z}q(Z|X,Y;\phi) \log\frac{p(\delta)p(Y,Z|X,\delta;\gamma)}{q(\delta)q(Z|X,Y;\phi)}\nonumber\\
    &= \int_{\delta} q(\delta) \int_{Z}q(Z|X,Y;\phi) \log\frac{p(Y,Z|X,\delta;\gamma)}{q(Z|X,Y;\phi)}\nonumber + \int_{\delta} q(\delta) \int_{Z}q(Z|X,Y;\phi) \log\frac{p(\delta)}{q(\delta)}\nonumber\\
    &= \mathbb{E}_{q(Z|X,Y;\phi)}\left[ \log\frac{p(Y,Z,\delta|X,\delta;\gamma)}{q(Z|X,Y;\phi)}\right] 
\end{align}

\section{Experiment Setup Details}
\label{sec:app_experiment_setup_details}

Here we describe additional details about experiment setup:

\begin{itemize}
    \itemsep0em
    \item In Section \ref{sec:swaps_metric}, we describe the swaps metric.
    \item In Section \ref{sec:sup_baseline}, we describe the specific parameters used for the baselines.
    \item In Section \ref{sec:sup_opthp}, we specify the optimal hyperparameters chosen for sigmoid experiments and clinical experiments.
    \item In Section \ref{sec:sup_clin_bio}, we detail the biomarkers and baseline features used for both clinical datasets: heart failure and Parkinson's disease.
    \item In Section \ref{sec:missing_values}, we describe how SubLign handles missing values, including irregularly lengths of time-series data.
    \item In Section \ref{sec:statistical_sig}, we describe how we compute statistical significance for results.
\end{itemize}  

\subsection{Swaps Metric\label{sec:swaps_metric}}

For true sorted alignment values $a_1, \ldots, a_N$ for $N$ patients, we define the swaps metric $\mathcal{S}$ of proposed alignment values $b_1, \ldots, b_N$ as the number of swaps needed to sort the predicted disease times into the true disease stages, expressed as percentage of total possible swaps. 

$$\mathcal{S} = \frac{\sum_{i,j; i<j} \mathbbm{1}(a_i < b_i, a_j > b_j)}{N (N-1) / 2}  $$





\subsection{Baselines\label{sec:sup_baseline}}

We outline SubNoLign and KMeans+Loss in the main paper. Here we describe the remaining baselines.

\paragraph{SuStaIn} SuStaIn~\cite{young2018uncovering} is a disease progression algorithm that recovers subtype and stage from cross-sectional data. We transform our longitudinal data by dropping patient affiliation across visits. We transform the data by subtracting the mean for each feature and dividing by the standard deviation for each feature. We assume the Z-scored values have a max of 5. We run for 1,000,000 epochs for the Markov Chain Monte Carlo sampling and 1,000 epochs for optimization. We use an open source implementation by the authors.\footnote{https://github.com/ucl-pond/pySuStaIn}

\paragraph{Bayesian approach} The Bayesian approach~\cite{huopaniemi2014disease} assumes longitudinal data, but there must be a small number of measured time points. We assume that there are 10 observed time points where observed data can begin as well as a window of 10 time points before the observed window where a patient's values can be aligned to. Because biomarker values are scaled between 0 and 10, we assume that values change between time points based on a Gaussian with $\sigma=2$ and that subtype means for each time point are drawn from a Gaussian with $\sigma=5$. We draw 4000 samples and use the maximum a posteriori estimate to determine stage and subtype for test patients. Because we could not find an open-source option, we implemented the algorithm ourselves based on the description in the paper.

\paragraph{PAGA} Partition-based graph abstraction, or PAGA,~\cite{wolf2019paga} assumes cross-sectional data, so we create separate visits for each patient visit. For algorithm parameters, we set resolution to 0.05, number of neighbors to 15, and connectivity cutoff of 0.05. We use an open source implementation by the authors.\footnote{https://github.com/dynverse/ti\_paga/blob/master/run.py}

\paragraph{Dynamic time warping} Dynamic time warping (DTW) defines similarity between time series that can be combined with clustering techniques. DTW methods include using soft-DTW~\cite{cuturi2017soft} and kernel~\cite{cuturi2011fast} before using K-means with the chosen similarity metric. We use open source implementations of DTW algorithms\footnote{https://pypi.org/project/dtw-python/} to generate our baseline comparisons. 



\paragraph{Tensor factorization} Sparse tensor factorization has been used for disease phenotyping. The decomposition of large and sparse datasets using canonical polyadic decomposition can create an interpretable output for phenotyping. We use the Matlab open source implementation of SPARTan. \footnote{https://github.com/kperros/SPARTan} We found these baseline results to yield poor clustering performance despite aggressive hyperparameter tuning.  We surmise this is because transforming our data from continuous to discrete time resulted in a very large and extremely sparse matrix factorizing which is a tricky optimization problem. 

\subsection{Optimal hyperparameters for sigmoid and clinical baselines\label{sec:sup_opthp}}

For all models, we run for 1000 epochs and use the model with the best training loss over the 1000 epochs for evaluation. For the sigmoid dataset, the optimal hyperparameters are latent space of dimension 5, 100 hidden units in the RNN, 50 hidden units in the multi-layer perceptron, learning rate of 0.01, and no regularization.



For the Parkinson's disease dataset, we searched on a slightly smaller set of hyperparameters for SubLign and found optimal hyperparameters of $\beta=0.01$, no regularization, 10 latent dimensions, 10 hidden units for the multi-layer perceptron, 200 units for the recurrent neural network, and learning rate of 0.1.

For the heart failure dataset, we searched on a slightly smaller set of hyperparameters for SubLign and found optimal hyperparameters of $\beta=0.001$, no regularization, 10 latent dimensions, 20 hidden units for the multi-layer perceptron, 50 units for the recurrent neural network, and learning rate of 0.01.

\subsection{Clinical dataset biomarkers and baseline features \label{sec:sup_clin_bio}}
The source of the heart failure dataset will be revealed after blind peer review. 
For the heart failure dataset, we include the following biomarkers: Aorta - Ascending,
Aorta - Valve Level,
Aortic Valve - Peak Velocity,
Left Atrium - Four Chamber Length,
Left Atrium - Long Axis Dimension,
Left Ventricle - Diastolic Dimension,
Left Ventricle - Ejection Fraction,
Left Ventricle - Inferolateral Thickness,
Left Ventricle - Septal Wall Thickness,
Mitral Valve - E Wave,
Mitral Valve - E Wave Deceleration Time,
and Right Atrium - Four Chamber Length.

From the Parkinson's Progression Markers Initiative (PPMI) dataset, we include four main biomarkers: 1) MOCA, a cognitive assessment, 2) SCOPA-AUT, an autonomic assessment, 3) NUPDRS1, an assessment of non-motor symptoms, and 4) a maximum taken over NUPDRS3 and NUPDRS2 as an assessment of motor symptoms. We removed patients without extractable biomarker measurements and include preprocessing code.


The baseline features considered for heart failure are: age, anemia, atherosclerosis, atrial fibrillation, Black, body mass index, chronic kidney disease, diastolic heart failure, esophageal reflux, female, hyperlipidemia, hypertension, hypothyroidism, kidney disease, major depressive disorder, obesity
old myocardial infarction, other race, pulmonary heart disease, pneumonia, renal failure, type 2 diabetes, urinary tract infection, and White.

The baseline features considered for Parkinson's disease (PD) are: male, Hispanic/Latino, White, Asian, Black, American Indian, Pacific Islander, not specified race, biological mom with PD, biological dad with PD, full sibling with PD, half sibling with PD, maternal grandparent with PD, paternal grandparent with PD, maternal aunt/uncle with PD, paternal aunt/uncle with PD, kids with PD, years of education, right handed, left handed, University of Pennsylvania Smell Identification Test (UPSIT) part 1, UPSIT part 2, UPSIT part 3, UPSIT part 4, and UPSIT total.

\subsection{Missing values}
\label{sec:missing_values}
SubLign allows for missing biomarker dimensions and missing patient visits to accommodate the sparsity of clinical data. For missing visits, we adapt the recognition network to handle variable sequence lengths. We mask out missing observations so they have no contribution to the learning stage, except for the recognition network input. For the recognition network input, we linearly interpolate missing values for each patient. For baselines that cannot handle missing data, we also linearly interpolate missing values for each patient.

\subsection{Statistical significance}
\label{sec:statistical_sig}
To estimate robustness of our models, we evaluate our held-out performance over 5 trials. Each trial consists of randomized 60/20/20 training/validation/test data folds and a different random seed. In order to compare models across 5 trials, we report the means and standard deviations from the 5 trials. When the reported performance intervals overlap, we compute the statistical significance of the pairwise differences using a t-test and a Benjamini-Hochberg correction of 0.05. 


\section{Additional Experiment Results}
\label{sec:app_add_experiment_results}
\begin{itemize}
    \itemsep0em
    \item In Section \ref{sec:sublign_sigmoid_viz}, we visualize the SubLign and SubNoLign subtypes for sigmoid data.
    \item In Section \ref{sec:model_misspecification}, we present results on model misspecification.
    \item In Section~\ref{sec:missing_experiments}, we present the empirical results with varying levels of missingness.
        
\end{itemize}

\subsection{SubLign and SubNoLign subtype visualization}
\label{sec:sublign_sigmoid_viz}

\begin{figure}[ht]
    \centering
    \includegraphics[width=0.35\textwidth]{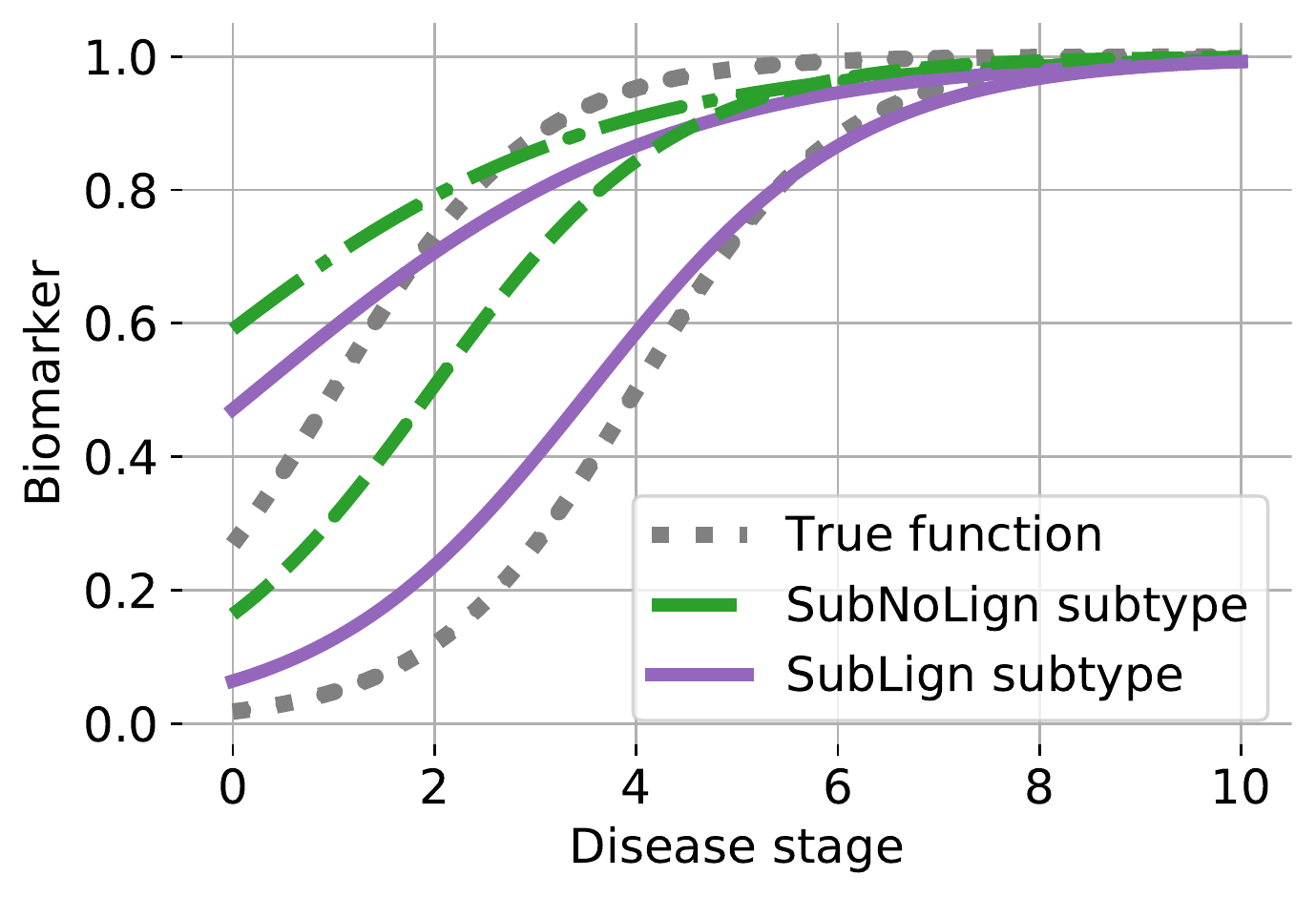}
    \caption{One of three dimensions of learned SubLign and SubNoLign subtypes from sigmoid synthetic data plotted on top of original data generating functions.}
    \label{fig:subnolign_viz}
\end{figure}

In Figure~\ref{fig:subnolign_viz}, we show the visualization for SubLign subtypes compared to SubNoLign for the first dimension of the sigmoid dataset. We find that the visualization of the SubNoLign subtypes are not as close to the data generating function as the SubLign subtypes. All other parameters, data dimensions, and experimental conditions are held constant.

\subsection{Model Misspecification}
\label{sec:model_misspecification}

Because our model implicitly assumes a functional form (e.g., sigmoid or quadratic), we investigate learning under model misspecification. With synthetic datasets created using splines on 5 randomly generated control points with the same noise rates, dimensions, and censoring as the original experiments, we create two settings: one where the control points are monotonically increasing (using SubLign with a sigmoid $f$ function) and one with no restrictions (using SubLign with a quadratic $f$ function).

We run SubLign and use a sigmoid $f$ function the monotonically increasing control points (labeled ``Incr") and use quadratic $f$ function for splines generated without restrictions (labeled ``Any").

In Table~\ref{table:misspec}, we present results on model misspecification over 2 settings and 5 trials of SubLign learning on data generated from splines with 5 control points over 3 dimensions. We conclude that SubLign is robust against reasonable model misspecification using piecewise polynomial data. 


\begin{table}[th]
\centering
\begin{tabular}{l c c c}
\toprule
  {\sc Model} & {\sc ARI$\uparrow$} & {\sc Pearson $\uparrow$} & {\sc Swaps $\downarrow$} \\
    \midrule
    SubLign-Incr & 0.82 $\pm$ 0.17  & 0.83 $\pm$ 0.08 & 0.14 $\pm$ 0.04 \\
    SubNoLign-Incr &  0.77 $\pm$ 0.10 & -- & -- \\
    KMeans+Loss-Incr &  0.58 $\pm$ 0.11 & 0.43 $\pm$ 0.09& 0.21 $\pm$ 0.06 \\ 
    \midrule
    SubLign-Any & 0.46 $\pm$ 0.12 & 0.67 $\pm$ 0.39 & 0.22 $\pm$ 0.14 \\
    SubNoLign-Any & 0.29 $\pm$ 0.10 & -- & -- \\
    KMeans+Loss-Any &  0.22 $\pm$ 0.07 & 0.23 $\pm$ 0.21& 0.48 $\pm$ 0.11 \\
    \bottomrule
\end{tabular}
\caption{Model misspecification experiment means and standard deviations using 5 cubic splines datasets.}
\label{table:misspec}
\end{table}

\subsection{Missingness Experiments}
\label{sec:missing_experiments}


Here we present experiments varying the amount of missingness in synthetic datasets. Following the Parkinson's disease dataset (PPMI) where 47-60\% of the biomarkers are missing with a maximum of 17 observations for each patient, we modify our synthetic sigmoid dataset and remove biomarkers uniformly randomly with different missingness rates and $M=17$.

Additionally we experiment with two missingness imputation methods for the baseline models. In addition to the linear interpolation used in the results presented in the main paper, we experiment with two other imputation methods: chained equations (MICE)~\cite{white2011multiple} and a multi-directional recurrent neural network (MRNN) designed for multivariate time-series~\cite{yoon2018estimating}.

We find that SubLign outperforms baselines across higher missingness rates although with no missing observations, some baselines are comparable.
In Table~\ref{tab:missingness50}, we show results with 50\% of the data missing uniformly random. In Table~\ref{tab:missingness25}, we show results with 25\% of the data missing uniformly random. In Table~\ref{tab:missingness0}, we show results with none of the data missing uniformly random. 


\begin{table*}[ht]
    \centering
    \caption{Experiments on synthetic data with 50\% of the data missing. Baselines include SuStaIn~\cite{young2018uncovering}, BayLong~\cite{huopaniemi2014disease}, PAGA~\cite{wolf2019paga}, SPARTan~\cite{perros2017spartan}, clustering using Soft-DTW~\cite{cuturi2011fast}, and clustering using Kernel-DTW~\cite{dhillon2004kernel}. Imputation methods include MICE~\cite{white2011multiple} and MRNN ~\cite{yoon2018estimating}.}
    \begin{tabular}{ll l l l}
    \toprule
        {\sc Imputation Method} & {\sc Model} &  {\sc ARI $\uparrow$} & {\sc Swaps $\downarrow$}  & {\sc Pearson $\uparrow$}  \\
        \midrule
        -- & SubLign & 0.813 $\pm$ 0.024 & 0.299 $\pm$ 0.019 & 0.613 $\pm$ 0.049 \\
        -- & SubNoLign & 0.789 $\pm$ 0.058 & -- & -- \\
        
        \midrule
        MICE & KMeans+Loss &  0.780 $\pm$ 0.046 & 0.327 $\pm$ 0.048 & 0.503 $\pm$ 0.018  \\
        & SuStaIn & 0.459 $\pm$ 0.010 &
0.243 $\pm$ 0.004 &
0.160 $\pm$ 0.030 \\
         & BayLong & 0.028 $\pm$ 0.003 & 0.480 $\pm$ 0.002 & 0.009 $\pm$ 0.003 \\
         & PAGA &  0.003 $\pm$ 0.002 & 0.494$\pm$ 0.028 & 0.034 $\pm$ 0.001 \\
          -- & Soft-DTW & 0.081 $\pm$ 0.004 & -- & -- \\
        -- & Kernel-DTW & 0.013 $\pm$ 0.002 & -- & -- \\
         & 
         SPARTan &  0.081 $\pm$ 0.013 & -- & -- \\
        \midrule
        MRNN & KMeans+Loss & 0.783 $\pm$ 0.071 & 0.321 $\pm$ 0.120 & 0.562 $\pm$ 0.042\\
        & SuStaIn & 0.450 $\pm$ 0.120 & 0.304 $\pm$ 0.120 & 0.434 $\pm$ 0.120  \\
         & BayLong &  0.028 $\pm$ 0.003 & 0.480 $\pm$ 0.002 & 0.009 $\pm$ 0.003 \\
         & PAGA & 0.004 $\pm$ 0.001 & 0.492$\pm$ 0.031 & 0.032 $\pm$ 0.003 \\
         & Soft-DTW & 0.094 $\pm$ 0.005 & -- & -- \\
         & Kernel-DTW & 0.000 $\pm$ 0.003 & -- & -- \\
         & SPARTan & 0.091 $\pm$ 0.005 & -- & -- \\
        \bottomrule
    \end{tabular}
    \label{tab:missingness50}
\end{table*}

\begin{table*}[ht]
    \centering
    \caption{Experiments on synthetic data with 25\% of the data missing. Baselines include SuStaIn~\cite{young2018uncovering}, BayLong~\cite{huopaniemi2014disease}, PAGA~\cite{wolf2019paga}, SPARTan~\cite{perros2017spartan}, clustering using Soft-DTW~\cite{cuturi2011fast}, and clustering using Kernel-DTW~\cite{dhillon2004kernel}. Imputation methods include MICE~\cite{white2011multiple} and MRNN ~\cite{yoon2018estimating}.}
    \begin{tabular}{ll l l l}
    \toprule
        {\sc Imputation Method} & {\sc Model} &  {\sc ARI $\uparrow$} & {\sc Swaps $\downarrow$}  & {\sc Pearson $\uparrow$}  \\
        \midrule
        -- & SubLign & 0.811 $\pm$ 0.406 & 0.309 $\pm$ 0.028 & 0.554 $\pm$ 0.070 \\
        -- & SubNoLign & 0.743 $\pm$ 0.040 & -- & --\\
        
        \midrule
        MICE & KMeans+Loss & 0.741 $\pm$ 0.041 & 0.332 $\pm$ 0.048 & 0.508 $\pm$ 0.014 \\
        & SuStaIn &  0.611 $\pm$ 0.406 & 0.309 $\pm$ 0.028 & 0.554 $\pm$ 0.070  \\
         & BayLong &  0.111 $\pm$ 0.010 & 0.464 $\pm$ 0.071 & 0.011 $\pm$ 0.058 \\
         & PAGA & 0.010 $\pm$ 0.003 & 0.433 $\pm$ 0.007 & 0.048 $\pm$ 0.030 \\
         & Soft-DTW & 0.151 $\pm$ 0.031 & -- & -- \\
         & Kernel-DTW & 0.002 $\pm$ 0.004 & -- & --\\
         & SPARTan & 0.168 $\pm$ 0.008 & -- & --  \\
        \midrule
        MRNN & KMeans+Loss & 0.653 $\pm$ 0.029 & 0.308 $\pm$ 0.358 & 0.497 $\pm$ 0.025 \\
         & SuStaIn &  0.615 $\pm$ 0.112 & 0.244 $\pm$ 0.004 & 0.577 $\pm$ 0.020 \\
         & BayLong & 0.108 $\pm$ 0.016 & 0.461 $\pm$ 0.005 & 0.010 $\pm$ 0.041 \\
         & PAGA & 0.011 $\pm$ 0.002 & 0.413 $\pm$ 0.005 & 0.031 $\pm$ 0.002 \\
         & Soft-DTW & 0.103 $\pm$ 0.011 & -- & -- \\
         & Kernel-DTW & 0.006 $\pm$ 0.004 & -- & -- \\
         & SPARTan & 0.171 $\pm$ 0.041 & -- & -- \\
        \bottomrule
    \end{tabular}
    \label{tab:missingness25}
\end{table*}

\begin{table*}[ht]
    \centering
    \caption{Experiments on synthetic data with 0\% of the data missing.}
    \begin{tabular}{l l l l}
    \toprule
         {\sc Model} &  {\sc ARI $\uparrow$} & {\sc Swaps $\downarrow$}  & {\sc Pearson $\uparrow$}  \\
        \midrule
        SubLign & 0.980 $\pm$ 0.000 & 0.273 $\pm$ 0.012 & 0.714 $\pm$ 0.022 \\
         SubNoLign & 0.809 $\pm$ 0.382 & -- & --\\
        KMeans+Loss & 0.980 $\pm$ 0.016 & 0.057 $\pm$ 0.112 & 0.480 $\pm$ 0.039 \\ 
        SuStaIn~\cite{young2018uncovering} & 0.765 $\pm$ 0.012 & 0.144 $\pm$ 0.004 & 0.477 $\pm$ 0.020\\ 
        BayLong~\cite{huopaniemi2014disease} &  0.201 $\pm$ 0.166 & 0.451 $\pm$ 0.050 & 0.011 $\pm$ 0.021 \\
        PAGA~\cite{wolf2019paga} & 0.251 $\pm$ 0.031 & 0.481 $\pm$ 0.003 & 0.015 $\pm$ 0.002 \\
        Soft-DTW~\cite{cuturi2011fast} & 0.974 $\pm$ 0.015 & -- & --\\
         Kernel-DTW~\cite{dhillon2004kernel} & 0.949 $\pm$ 0.040 & -- & --\\
        SPARTan~\cite{perros2017spartan} & 0.251 $\pm$ 0.031 & -- & -- \\
        
        \bottomrule
    \end{tabular}
    \label{tab:missingness0}
\end{table*}

\section{Quadratic Data Results}
\label{sec:quadratic_data_results}
We describe an additional set of experiments using the quadratic functional family. These experiments were designed to better understand where SubLign is able to learn clustering and alignment metrics well.
\begin{itemize}
    \itemsep0em
    \item In Section~\ref{sec:quad_setup}, we detail the dataset creation.
    \item In Section~\ref{sec:quad_params}, we outline the optimal hyperparameters for the quadratic experiments.
    \item In Section~\ref{sec:quad_results}, we describe the empirical results for the quadratic experiments.
\end{itemize}
\subsection{Setup}
\label{sec:quad_setup}
For the quadratic dataset, we generate data from 2 subtypes and 1 dimension with generating functions. See Table~\ref{tab:quad_experiments} for subtype generating functions. Similar to the sigmoid synthetic dataset, for each patient in the datasets, we draw subtype $\kappa \sim Bern(0.5)$. The true disease stage is drawn $t_m \sim U(0,T^+)$ for observation $m \in [M]$. The biomarker values are drawn $y_m \sim N(\mu_{m}, S)$ where $\mu_{m} = \sum_{k \in [K]} \mathbbm{1}(\kappa = k) f_k(t_m)$. The observed disease time $x_m$ is shifted such that the first patient observation is at time 0. Therefore $x_m = t_m - \tau$ where $\tau = \min_{j \in [M]}  t_j$ and is the earliest true disease time for the patient. We sample $N=1000$ patients with $M=4$ patient observations times with noise $S=0.25$ and upper time bound $T^+ = 10$.

\begin{table*}[ht]
    \centering
    \vskip 0.15in
    \begin{tabular}{c l l}
    \toprule
        {\sc Figure} & {\sc Description} & {\sc Subtype Generating Functions}\\
        \midrule
        6 & Quadratic curve and flat line, separable & $f_1(t) = 0.25t^2 - 2.2t + 5,$\\ && $ f_2(t) = 2$ \\ 
        \midrule
        7 & Quadratic curve and flat line, overlapping & $f_1(t) = 0.25t^2 - 2.2t + 5,$\\ && $ f_2(t) = -2$ \\ 
        \midrule
        8 & Quadratic curve and sloped line, separable & $f_1(t) = 0.25t^2 - 2.2t + 5,$\\ && $ f_2(t) = 0.4t$ \\ 
        \midrule
        9 & Quadratic curve and sloped line, overlapping & $f_1(t) = 0.25t^2 - 2.2t + 5, $\\ && $f_2(t) = 0.4t - 5$ \\ 
        \midrule
        10 & Quadratic curves in opposite directions, separable & $f_1(t) = 0.25t^2 - 2.2t + 3,$\\ && $ f_2(t) = -0.25t^2 +2.2 -5$ \\ 
        \midrule
        11 &Quadratic curves in opposite directions, overlapping &  $f_1(t) = 0.25t^2 - 2.2t + 7,$\\ && $ f_2(t) = -0.25t^2 +2.2 -5$ \\ 
        \bottomrule
    \end{tabular}
    \caption{Quadratic dataset subtype generating functions and corresponding figure numbers}
    \label{tab:quad_experiments}
    \vskip -0.1in
\end{table*}

We construct our quadratic experiments such that we examine different model classes (i.e. flat, linear, quadratic) as well as examine subtypes that are overlapping or separable.

We include baseline results for the quadratic datasets. Note that SuStaIn~\cite{young2018uncovering} assumes monotonically increasing functions and is therefore omitted. We denote degenerate solutions with dashes.

\subsection{Optimal hyperparameters for quadratic datasets \label{sec:quad_params}}

For the synthetic quadratic dataset corresponding to Figure~\ref{fig:data3}, we found the optimal hyperparameters for SubLign as no regularization, 5 hidden dimensions for the multi-layer perceptron, 200 latent dimensions, 200 units for the recurrent neural network, and learning rate of 0.001.

For the synthetic quadratic dataset corresponding to Figure~\ref{fig:data4}, we found the optimal hyperparameters for SubLign as no regularization, 5 hidden dimensions for the multi-layer perceptron, 200 latent dimensions, 200 units for the recurrent neural network, and learning rate of 0.001.

For the synthetic quadratic dataset corresponding to Figure~\ref{fig:data5}, we found the optimal hyperparameters for SubLign as no regularization, 5 hidden dimensions for the multi-layer perceptron, 200 latent dimensions, 200 units for the recurrent neural network, and learning rate of 0.001.

For the synthetic quadratic dataset corresponding to Figure~\ref{fig:data6}, we found the optimal hyperparameters for SubLign as no regularization, 5 hidden dimensions for the multi-layer perceptron, 200 latent dimensions, 200 units for the recurrent neural network, and learning rate of 0.001.

For the synthetic quadratic dataset corresponding to Figure~\ref{fig:data7}, we found the optimal hyperparameters for SubLign as no regularization, 5 hidden dimensions for the multi-layer perceptron, 200 latent dimensions, 200 units for the recurrent neural network, and learning rate of 0.001.

For the synthetic quadratic dataset corresponding to Figure~\ref{fig:data7}, the optimal hyperparameters are latent space of dimension 10, 20 hidden units in the RNN, 50 hidden units in the multi-layer perceptron, learning rate of 0.01, and no regularization.

For the synthetic quadratic dataset corresponding to Figure~\ref{fig:data8}, the optimal hyperparameters are latent space of dimension 5, 100 hidden units in the RNN, 50 hidden units in the multi-layer perceptron, learning rate of 0.01, and no regularization.

\subsection{Results}
\label{sec:quad_results}
\begin{figure*}[h]
    \centering
    \includegraphics[width=0.3\textwidth]{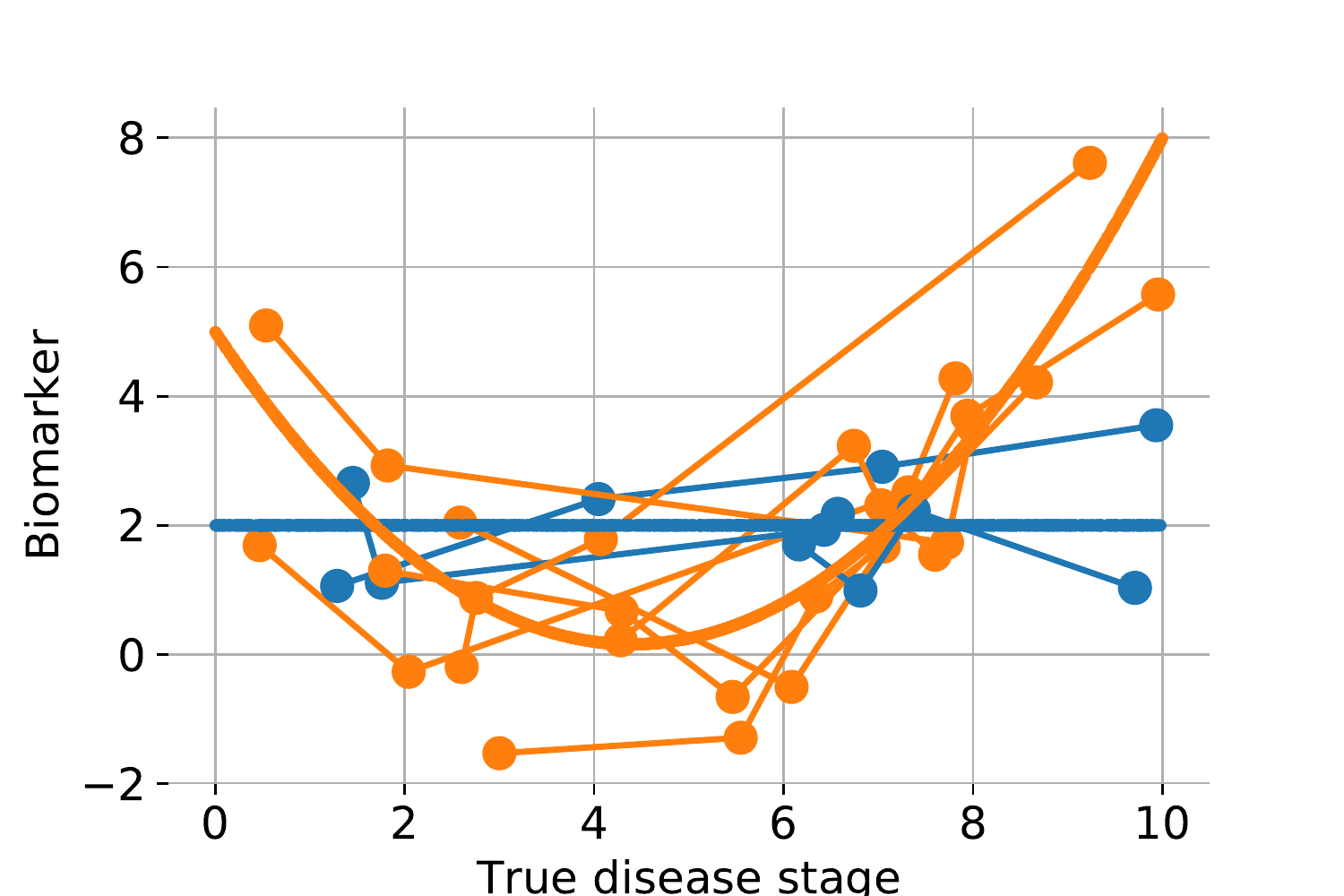}
    \vskip 0.15in
    \begin{tabular}{l r r r}
    \toprule
    {\sc Model} & {\sc ARI $\uparrow$} & {\sc Swaps $\downarrow$} & {\sc Pearson $\uparrow$} \\
    \midrule
SubLign &   0.277 $\pm$ 0.040 & 0.277 $\pm$ 0.014 & 0.516 $\pm$ 0.007\\
SubNoLign & 0.103 $\pm$ 0.002 & -- & --\\
KMeans+Loss & 0.213 $\pm$ 0.009 & 0.498 $\pm$ 0.022 & 0.016 $\pm$ 0.051\\
SuStaIn~\cite{young2018uncovering} & 0.151 &  0.203 &  0.000 \\
Bayesian~\cite{huopaniemi2014disease} & 0.000 $\pm$ 0.000 & 0.501 $\pm$ 0.017 & 0.018 $\pm$ 0.125\\
PAGA~\cite{wolf2019paga} & 0.027 $\pm$ 0.001 & -- & --\\
\bottomrule
    \end{tabular}
    \caption{Synthetic results over 5 trials. \textbf{Top}: Data generating functions for two subtypes (thick lines) and example aligned patients (dots and thin lines). \textbf{Bottom}: SubLign outperforms baselines while KMeans+Loss recovers subtypes (ARI metric) better than SubNoLign, but alignment metrics are difficult to recover because of the horizontal subtype}
    \label{fig:data3}
    \vskip -0.1in
\end{figure*}

\begin{figure*}
    \centering
    \includegraphics[width=0.3\textwidth]{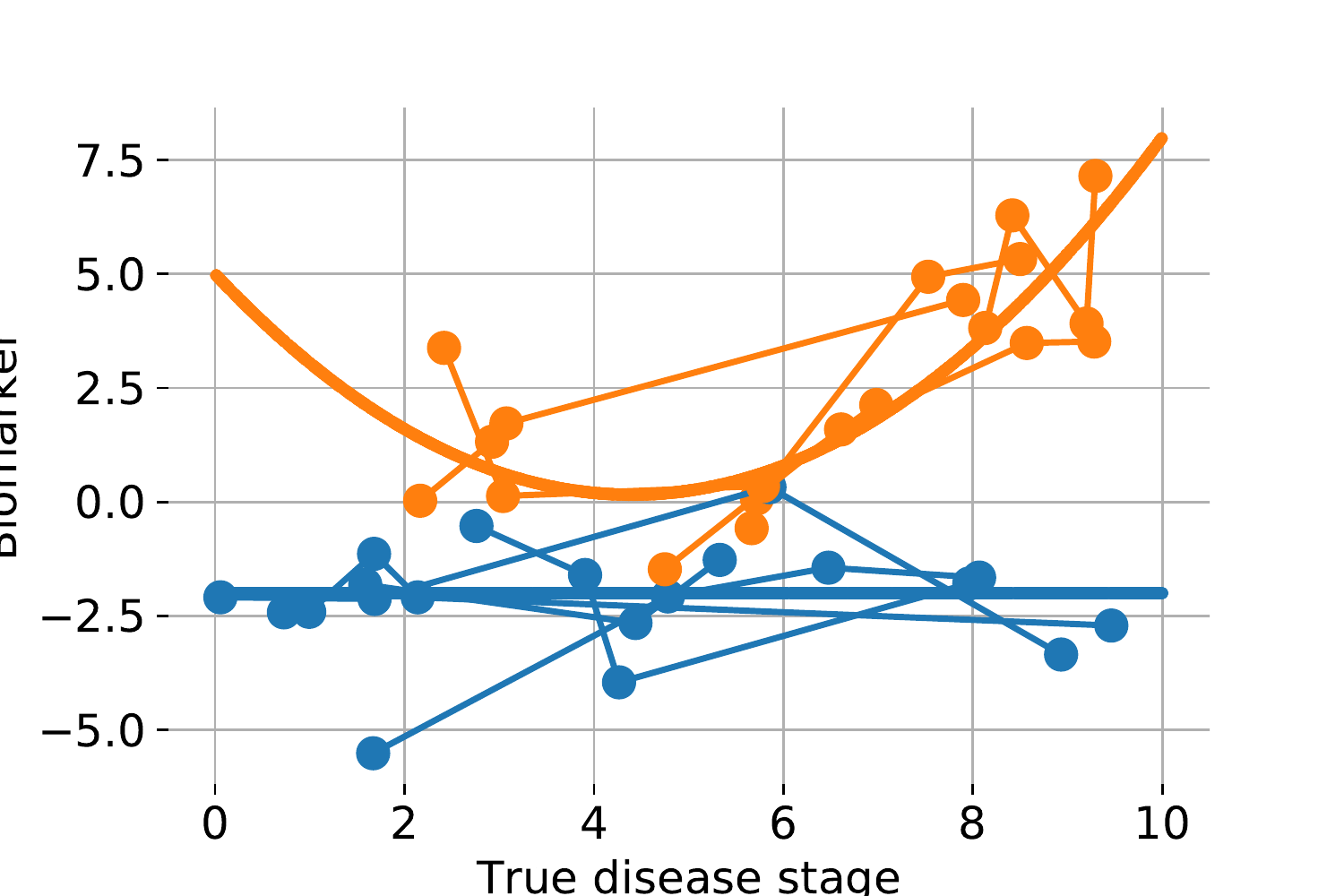}
    \vskip 0.15in
    \begin{tabular}{l r r r}
    \toprule
    \textbf{MODEL} & \textbf{ARI $\uparrow$} & \textbf{SWAPS $\downarrow$} & \textbf{PEARSON $\uparrow$} \\
    \midrule
SubLign &  0.980 $\pm$ 0.000 & 0.253 $\pm$ 0.001 & 0.527 $\pm$ 0.011\\
SubNoLign & 0.980 $\pm$ 0.000 & -- & -- \\
KMeans+Loss &  0.883 $\pm$ 0.000 & 0.471 $\pm$ 0.011 & 0.064 $\pm$ 0.067\\
SuStaIn~\cite{young2018uncovering} &  0.228 $\pm$ 0.039 & 0.182 $\pm$ 0.010 & 0.000 $\pm$ 0.000\\
Bayesian~\cite{huopaniemi2014disease} & 0.198 $\pm$ 0.189 & 0.446 $\pm$ 0.052 & 0.157 $\pm$ 0.286\\
PAGA~\cite{wolf2019paga} & 0.227 $\pm$ 0.035 & -- & -- \\
\bottomrule
    \end{tabular}
    \vskip -0.1in
    \caption{Synthetic results over 5 trials. \textbf{Top}: Data generating functions for two subtypes (thick lines) and example aligned patients (dots and thin lines). \textbf{Bottom}: SubLign and SubNoLign have near-perfect clustering accuracy (ARI) while alignment metrics (swaps, Pearson) are difficult recover because of the horizontal subtype.}
    \label{fig:data4}
\end{figure*}

\begin{figure*}
    \centering
    \includegraphics[width=0.3\textwidth]{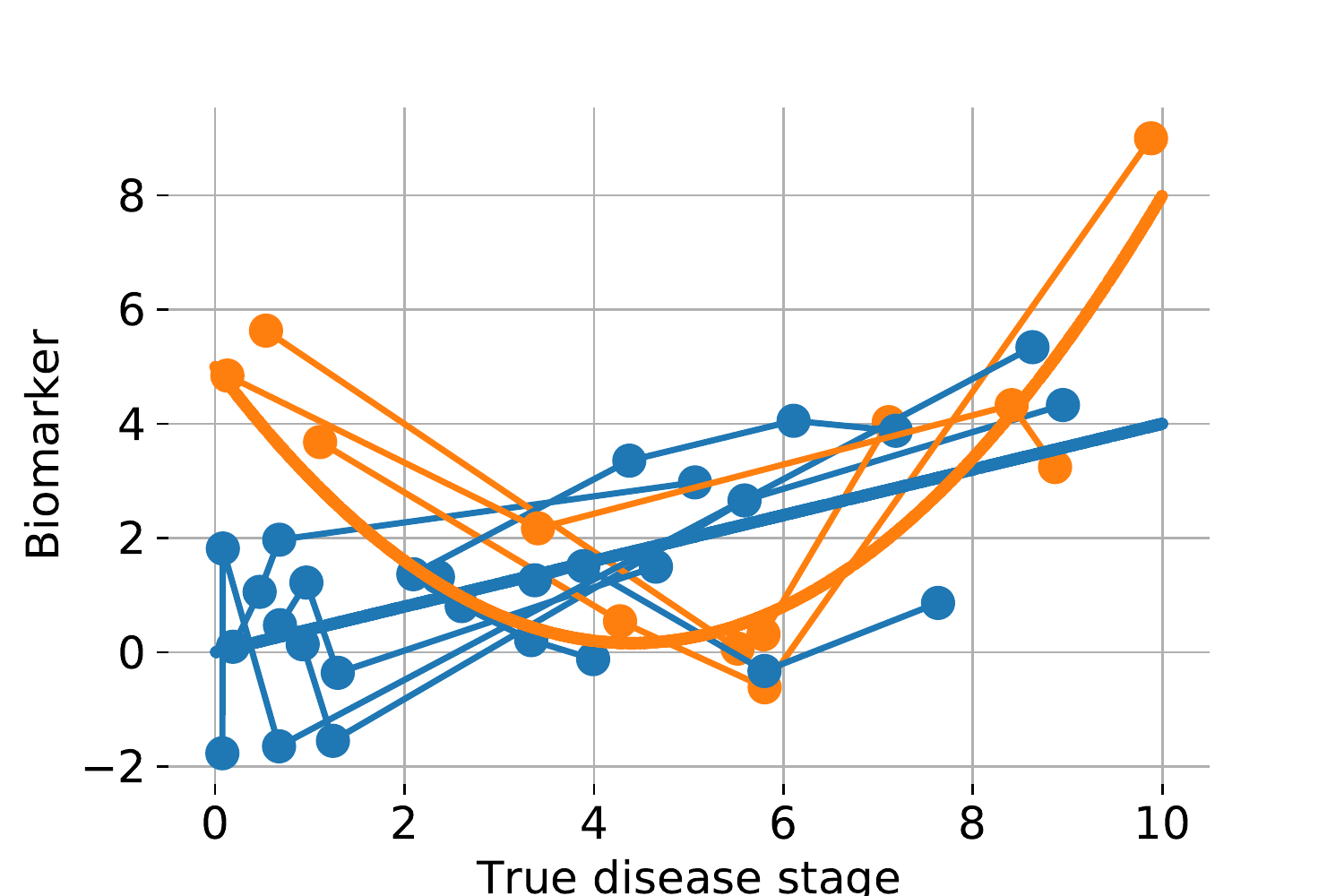}
    \vskip 0.15in
    \begin{tabular}{l r r r}
    \toprule
    {\sc Model} &  {\sc ARI $\uparrow$} & {\sc Swaps $\downarrow$} & {\sc Pearson $\uparrow$} \\
    \midrule
SubLign & 0.122 $\pm$ 0.006 & 0.272 $\pm$ 0.001 & 0.621 $\pm$ 0.020\\
SubNoLign &  0.145 $\pm$ 0.006 & -- & --\\
KMeans+Loss & 0.031 $\pm$ 0.030 & 0.302 $\pm$ 0.026 & 0.498 $\pm$ 0.010\\
SuStaIn~\cite{young2018uncovering} &  0.138 $\pm$ 0.019 & 0.119 $\pm$ 0.006 & 0.000 $\pm$ 0.000\\
Bayesian~\cite{huopaniemi2014disease} & 0.001 $\pm$ 0.002 & -- & -- \\
PAGA~\cite{wolf2019paga} & 0.009 $\pm$ 0.000 & -- & -- \\
\bottomrule
\end{tabular}
\vskip -0.1in
    \caption{Synthetic results over 5 trials. \textbf{Top}: Data generating functions for two subtypes (thick lines) and example aligned patients (dots and thin lines). \textbf{Bottom}: SubLign outperforms baselines in clustering and alignment metrics although the task is challenging.}
    \label{fig:data5}
\end{figure*}

\begin{figure*}
    \centering
    \includegraphics[width=0.3\textwidth]{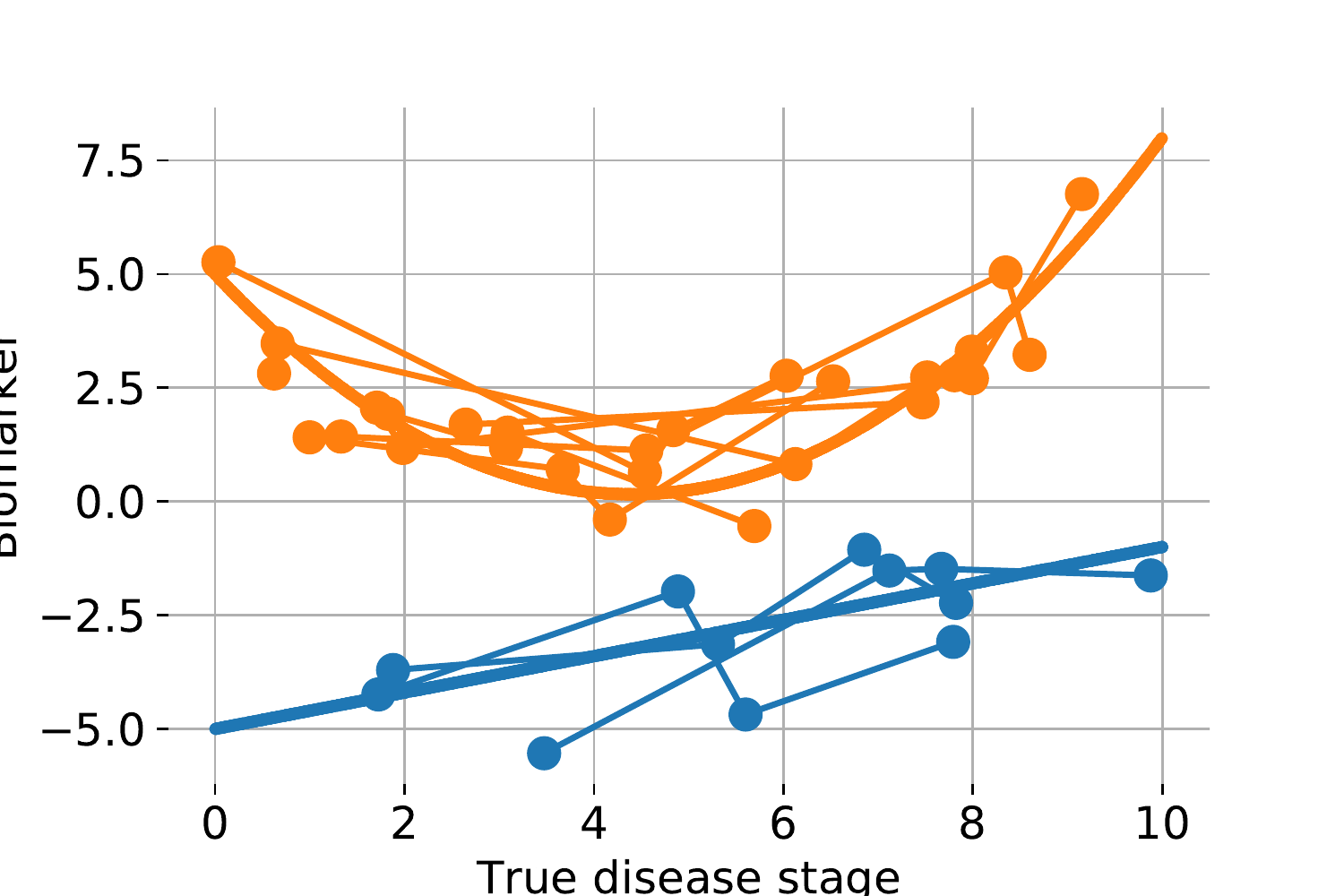}
    \vskip 0.15in
    \begin{tabular}{l r r r}
    \toprule
    {\sc Model} &  {ARI $\uparrow$} & {\sc Swaps $\downarrow$} & {\sc Pearson $\uparrow$} \\
    \midrule
SubLign & 0.968 $\pm$ 0.016 & 0.253 $\pm$ 0.015 & 0.604 $\pm$ 0.041\\
SubNoLign &  0.968 $\pm$ 0.016 & -- & -- \\
KMeans+Loss &  0.964 $\pm$ 0.008 & 0.486 $\pm$ 0.032 & 0.044 $\pm$ 0.101 \\
SuStaIn~\cite{young2018uncovering} & 0.220 $\pm$ 0.011 & 0.196 $\pm$ 0.008 & 0.000 $\pm$ 0.000\\
Bayesian~\cite{huopaniemi2014disease} & 0.221 $\pm$ 0.214 & 0.448 $\pm$ 0.070 & 0.165 $\pm$ 0.193\\
PAGA~\cite{wolf2019paga} & 0.205 $\pm$ 0.012 & -- & --\\
\bottomrule
    \end{tabular}
    \vskip -0.1in
    \caption{Synthetic results over 5 trials. \textbf{Top}: Data generating functions for two subtypes (thick lines) and example aligned patients (dots and thin lines). \textbf{Bottom}: SubLign, SubNoLign, and KMeans+Loss perform well on clustering.}
    \label{fig:data6}
\end{figure*}

\begin{figure*}
    \centering
    \includegraphics[width=0.3\textwidth]{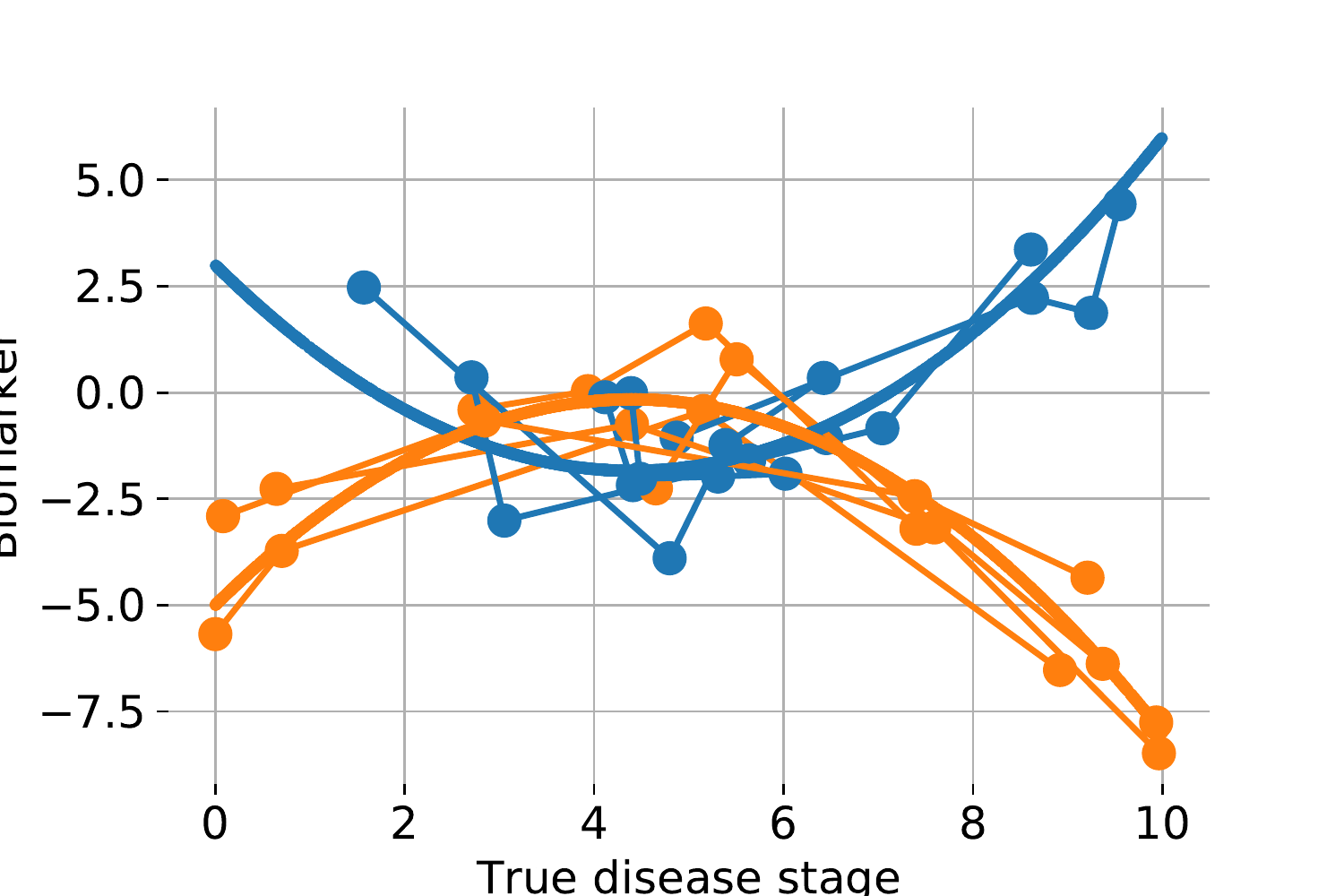}
    \vskip 0.15in
    \begin{tabular}{l r r r}
    \toprule
    {\sc Model} &  {\sc ARI $\uparrow$} & {\sc Swaps $\downarrow$} & {\sc Pearson $\uparrow$} \\
    \midrule
    SubLign & 0.729 $\pm$ 0.049 & 0.153 $\pm$ 0.006 & 0.843 $\pm$ 0.016\\
    SubNoLign & 0.721 $\pm$ 0.035 & -- & -- \\
KMeans+Loss & 0.540 $\pm$ 0.034 &  0.490 $\pm$ 0.020 & 0.043 $\pm$ 0.057\\
SuStaIn~\cite{young2018uncovering} &0.198 $\pm$ 0.024 & 0.200 $\pm$ 0.008 & 0.000 $\pm$ 0.000\\
Bayesian~\cite{huopaniemi2014disease} & 0.003 $\pm$ 0.007 & -- & -- \\
PAGA~\cite{wolf2019paga} & 0.059 $\pm$ 0.006 & -- & -- \\
\bottomrule
    \end{tabular}
    \vskip -0.1in
    \caption{Synthetic results over 5 trials. \textbf{Top}: Data generating functions for two subtypes (thick lines) and example aligned patients (dots and thin lines). \textbf{Bottom}: SubLign learns subtypes and recovers alignment better than baselines.}
    \label{fig:data7}
\end{figure*}

\begin{figure*}
    \centering
    \includegraphics[width=0.3\textwidth]{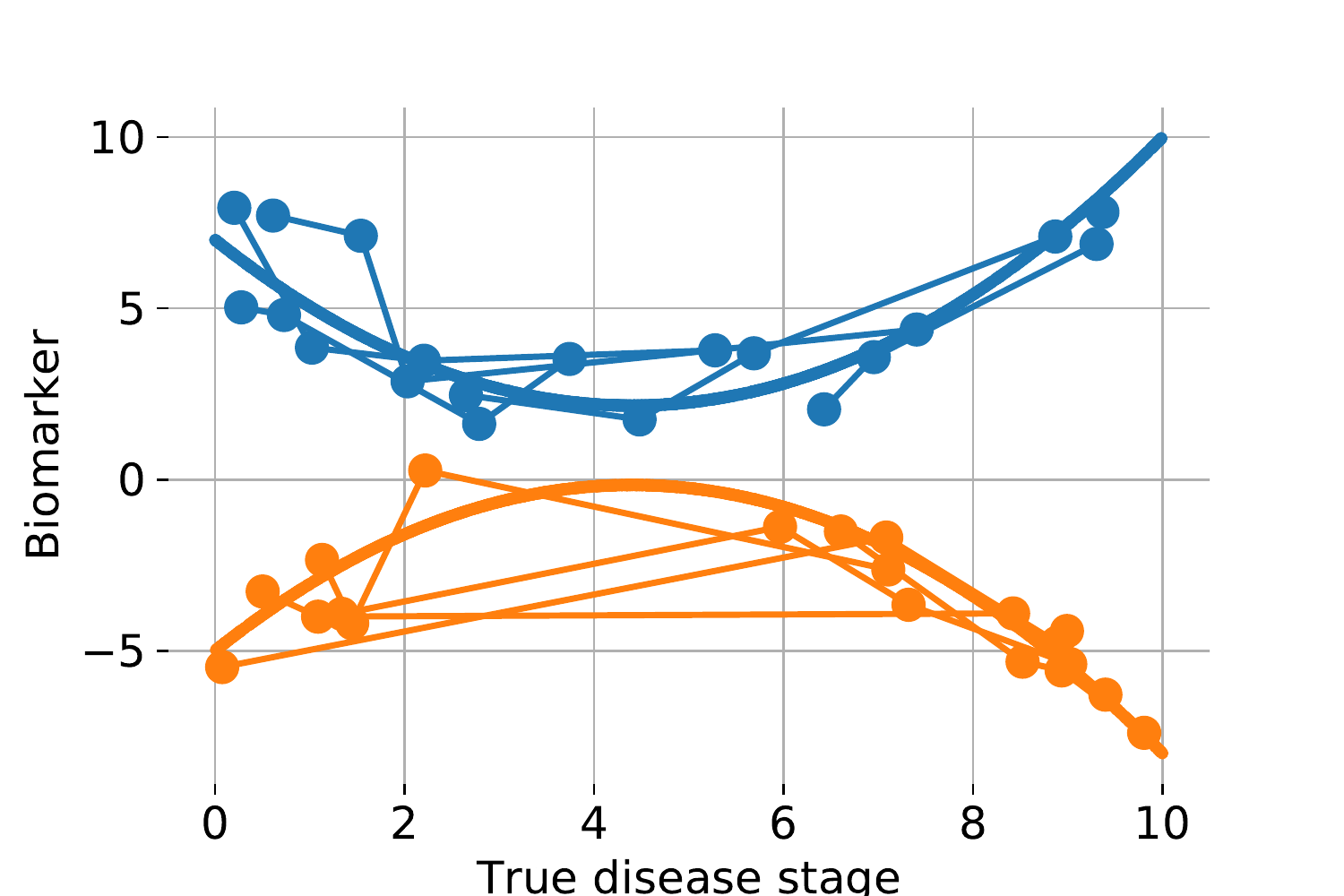}
    \vskip 0.15in
    \begin{tabular}{l r r r}
    \toprule
    {\sc Model} &  {\sc ARI $\uparrow$} & {\sc Swaps $\downarrow$} & {\sc Pearson $\uparrow$} \\
    \midrule
    
    SubLign & 1.000 $\pm$ 0.000 & 0.134 $\pm$ 0.012 & 0.907 $\pm$ 0.009\\
SubNoLign & 0.984 $\pm$ 0.008 & -- & --\\
KMeans+Loss & 1.000 $\pm$ 0.000 & 0.508 $\pm$ 0.034 & 0.014 $\pm$ 0.095 \\
SuStaIn~\cite{young2018uncovering} & 0.148 $\pm$ 0.032 & 0.247 $\pm$ 0.016 & 0.000 $\pm$ 0.000\\
Bayesian~\cite{huopaniemi2014disease} & 0.261 $\pm$ 0.349 & 0.409 $\pm$ 0.052 & 0.303 $\pm$ 0.119 \\
PAGA~\cite{wolf2019paga} & 0.233 $\pm$ 0.017 & -- & --\\
\bottomrule
\end{tabular}   
\vskip -0.1in
    \caption{Synthetic results over 5 trials. \textbf{Top}: Data generating functions for two subtypes (thick lines) and example aligned patients (dots and thin lines). \textbf{Bottom}: SubLign learns subtypes and alignment values.}
    \label{fig:data8}
\end{figure*}

In Figures~\ref{fig:data3} to~\ref{fig:data8}, we present the quantitative results of 6 different quadratic cases as well as a plot of example data and the data generating subtypes. In each, we see that SubLign or SubNoLign outperforms the baselines. 

When the subtypes are separable (i.e. Fig~\ref{fig:data4},~\ref{fig:data6}, and~\ref{fig:data8}), SubLign handily recovers the subtypes. When the subtypes are not separable (i.e. Fig~\ref{fig:data3},~\ref{fig:data5}, and~\ref{fig:data7}), SubLign still outperforms baselines. 

We note that alignment metrics are especially challenging to recover when one subtype is a flat or sloped line as in with Figure~\ref{fig:data3} to \ref{fig:data8}. Because the alignment metric is entirely unidentifiable, the swaps and Pearson metrics suffer. Note that for the swaps metric, 0.5 corresponds to random guessing, so the lack of identifiability of one of the subtypes would cause a swaps metric of 0.25. When the second subtype has a changing slope, as in Figure~\ref{fig:data7}, the alignment metrics are more recoverable.

When the model is degenerate and does not return the alignment values, we denote this with an empty cell.


\section{Data Privacy and Ethics}
\label{sec:data_privacy_ethics}

The heart failure dataset was collected from a large health system in the United States and shared with us under a research data use agreement. The hospital obtained the relevant consent from individuals to use their data, and the dataset is covered by the hospital’s institutional review board (IRB). Since the dataset was de-identified and provided with a data use agreement, our institution’s IRB ruled it as exempt. The dataset use agreement was approved by the legal teams from both our academic institution and the hospital. 

The Parkinson’s Progression Markers Initiative (PPMI) provides open and full access to the study data, which is intended for researchers to study the disease. The PPMI dataset only includes patients who consent to including their data in the study, and the patients are de-identified in the dataset. 

\section{SubLign Clinical Insights}
\label{sec:app_clinical_insights}
In Figure~\ref{fig:hf}, we show the statistically significant baseline features from our SubLign results on heart failure patients. In Figure~\ref{fig:ppmi}, we show the statistically significant baseline features for our SubLign results on both Parkinson's disease patients and healthy controls as well as only Parkinson's disease patients.

\begin{figure*}[h]
\centering
\subfloat[width=0.45\textwidth][]{
\begin{tabular}{l r r}
\toprule
{\sc Feature} & A (321)  & B (298)\\
\midrule
Biological Dad With PD & 0.02 & 0.06 \\ 
Sibling With PD & 0.01 & 0.05 \\ 
UPSIT Part 1 & 7.55 & 5.49 \\ 
UPSIT Part 2 & 7.64 & 5.69 \\ 
UPSIT Part 3 & 6.98 & 5.23 \\ 
UPSIT Part 4 & 7.53 & 5.62 \\ 
UPSIT Total & 29.73 & 22.05 \\ 
\bottomrule
\end{tabular}
}
\subfloat[width=0.45\textwidth][]{
\begin{tabular}{l r r r}
\toprule
    {\sc Feature} & A (156) & B (112) & B (155)\\ 
\midrule
Male & $0.51 $ & $0.68 $ & $0.79 $ \\ 
White & $0.98 $ & $0.95 $ & $0.92 $ \\ 
UPSIT Part 1 & $6.01 $ & $5.07 $ & $5.43 $ \\ 
UPSIT Part 2 & $6.65 $ & $5.30 $ & $5.52 $ \\ 
UPSIT Part 3 & $5.92 $ & $4.79 $ & $4.90 $ \\ 
UPSIT Part 4 & $6.28 $ & $5.20 $ & $5.42 $ \\ 
UPSIT Total & $24.87 $ & $20.36 $ & $21.26 $ \\
\bottomrule
\end{tabular}
}
\caption{Subtypes found by SubLign for (a) 619 Parkinson's disease patients and healthy controls and (b) 423 Parkinson's disease patients only. Only statistically significant means between subtypes according to an ANOVA test with $p<0.05$ with a Benjamini-Hochberg correction are listed.}
\label{fig:ppmi}
\end{figure*}
    
\begin{table*}[]
    \centering
        \begin{tabular}{l r r}
\toprule
{\sc Feature} & A (321)  & B (298)\\
\midrule
Biological Dad With PD & 0.028 & 0.068 \\ 
Sibling With PD & 0.010 & 0.058 \\ 
UPSIT Part 1 & 7.558 & 5.493 \\ 
UPSIT Part 2 & 7.648 & 5.695 \\ 
UPSIT Part 3 & 6.988 & 5.238 \\ 
UPSIT Part 4 & 7.539 & 5.624 \\ 
UPSIT Total & 29.73 & 22.05 \\ 
\bottomrule
\end{tabular}
    \caption{Subtypes found by SubLign from Parkinson's disease patients and healthy controls using sparsely collected biomarkers. Only statistically significant means between subtypes according to an ANOVA test with $p<0.05$ are listed.}
    \label{tab:my_label}
\end{table*}
\begin{figure*}[h]
    \centering
    \begin{small}
    \begin{tabular}{l r r r}
    \toprule
    {\sc Feature} & A (674)  & B (444) & C (416)\\
    \midrule
    Age & 75.98 & 74.73 & 69.43 \\ 
    Female & 0.71 & 0.23 & 0.43 \\ 
    Anemia & 0.23 & 0.16 & 0.14 \\ 
    Atherosclerosis & 0.28 & 0.34 & 0.40 \\ 
    Atrial Fibrillation & 0.44 & 0.55 & 0.43 \\ 
    Chronic Kidney Disease & 0.27 & 0.34 & 0.34 \\ 
    Diastolic Heart Failure & 0.50 & 0.36 & 0.06 \\ 
    Obese & 0.56 & 0.65 & 0.46 \\ 
    Old Myocardial Infarction & 0.12 & 0.14 & 0.24 \\ 
    Pulmonary Heart Disease & 0.29 & 0.22 & 0.19 \\ 
    Systolic HF & 0.09 & 0.27 & 0.53 \\
    \bottomrule
    \end{tabular}
    \end{small}
    
    \caption{Subtypes found by SubLign from heart failure patients using echocardiogram biomarkers. Only statistically significant means between subtypes according to an ANOVA test with $p<0.05$ with a Benjamini-Hochberg correction are listed.}
    \label{fig:hf}
\end{figure*}

In comparison, we show the KMeans+Loss subtypes on the heart failure dataset (Table~\ref{tab:hf_kmeans}).

\begin{table*}[h!]
\centering
\begin{tabular}{l r r r}
\toprule
{\sc Feature} & { \sc A (240)} & {\sc B (802)} & \textbf{\sc C (492)}\\ 
\midrule 
Age & 71.567 & 74.565 & 73.793 \\ 
Hyperlipidemia & 0.529 & 0.448 & 0.541 \\ 
Chronic Kidney Disease & 0.346 & 0.273 & 0.370 \\ 
Esophageal Reflux & 0.375 & 0.259 & 0.289 \\ 
Pulmonary Heart Disease & 0.367 & 0.204 & 0.256 \\ 
Kidney Disease & 0.254 & 0.200 & 0.278 \\ 
Atherosclerosis & 0.196 & 0.131 & 0.201 \\ 
Anemia & 0.217 & 0.163 & 0.213 \\ 
Obese & 0.688 & 0.500 & 0.608 \\ 
\bottomrule
\end{tabular}
\caption{Heart Failure KMeans+Loss subtypes (patient counts in parentheses), described by mean baseline features. Only statistically significant features are listed and do not include systolic and diastolic HF, two known phenotypes of HF.}
\label{tab:hf_kmeans}
\end{table*}

%% file: identifiability_proof_2.tex
\begin{algorithm}[ht]
   \caption{Procedure for the identification of model parameters}\label{alg:identifiability}
\begin{algorithmic}[1]
   \STATE {\bfseries Input:} Observation times $X \in \mathbb{R}^{N \times M}$, biomarkers $Y \in \mathbb{R}^{N \times M \times D}$, polynomial degree $P$, invertible function $f$
    \STATE{\bfseries Output:} $\theta^P$, $\delta_1,\ldots,\delta_N$, $s_1,\ldots,s_N$ for each patient
    \STATE \textbf{Step 1: Transform the observed biomarkers;} $Q=f^{-1}(Y)$
    \STATE \textbf{Step 2: Obtain time-shifts using a single biomarker;}
    \STATE a) For each patient $i$, estimate the parameters $\dot{\theta^1}_i$ of $\kappa(x;\dot{\theta^1}_i)$ using a single biomarker $((x_{i,1},q_{i,1}),\ldots,(x_{i,M},q_{i,M})$ via polynomial regression, 
    \STATE b) Compute up to $P$ roots of polynomial $\kappa(x, \dot{\theta^1}_i)$ for each patient $i$ as $R_i=\{r_1,\ldots,r_P\}$ and set $\xi_i=\min \text{Real}(R_i)$ where $\text{Real}$ denotes the real part of (potentially complex) roots. 
    \STATE c) Estimate $\tilde{\theta^1}_i$ for polynomials in a \emph{canonical position} using $((x_{i,1}-\xi_i,q_{i,1}),\ldots,(x_{i,M}-\xi_i,q_{i,M})$ via polynomial regression,
    \STATE d) Cluster $\tilde{\theta^j}_i$ across patients via K-means clustering to yield cluster identities $s_1,\ldots,s_N$
    \STATE e) $\forall k,\;\; \eta_k = \min \{\xi_i \, | \, i \text{ s.t. } s_i=k\}$ and $\forall i,\;\;\delta_i = \xi_i-\eta_{s_i}$
    \STATE \textbf{Step 3: Estimate true polynomial coefficients using shifted observation times;}
    \FOR{biomarker $j=1$ {\bfseries to} $J$}
    \STATE For each patient, estimate the parameters $\theta^j_i$ of $\kappa(x;\hat{\theta^j}_i)$ using $((x_{1,1}-\delta_i,q_{1,1}[j]),\ldots,(x_{1,M}-\delta_i,q_{1,M}[j])$ via polynomial regression, 
    \ENDFOR
    \STATE Return $\theta^P=[\theta^1|\ldots|\theta^J],\{\delta_1,\ldots, \delta_N\},\{s_1,\ldots, s_N\}$
\end{algorithmic}
\end{algorithm}

We restate our assumptions. 

\begin{assumption_supplement}
$f$ is invertible, and $\kappa(x,\theta) = \theta_{0} +\sum_{p=1}^P \theta_p x^{p}$ describes a family of polynomial functions in x with parameter $\theta$ and degree $P>0$. The parameters of each subtype are unique.
\label{assum_suppl:functionals}
\end{assumption_supplement}


\begin{assumption_supplement}
$M\geq P+1$, \, i.e., for each object, across all the $D$ features, we observe at least $P+1$ values. 
\label{assum_suppl:enough_obs}
\end{assumption_supplement}

\begin{assumption_supplement}
For each subtype $s_k$, there exists an object $i$ whose alignment $\delta_i=0$.
\label{assum_suppl:delta0}
\end{assumption_supplement}

We provide the proof for Theorem \ref{thm:identifiability} below: 

\begin{proof}
The proof is constructive; i.e. we give an algorithm for the identification of the parameters of the model in Equation \ref{eqn:identifiable_model}. 
The algorithm for identification is presented in Algorithm \ref{alg:identifiability} and proceeds in three steps.

\textbf{Step 1:} The first step transforms the observed biomarkers by applying the inverse of function $f$, which exists by Assumption \ref{assum:functionals}. This leaves us with data as: $$f^{-1}(y_{i,m}) = \kappa(x_{i,m} + \delta_i; \theta^P_{s_i})\; \;\forall i \in N, m \in M$$
i.e. for all bio-markers, across all patients, we have data arising from different polynomial functions. 

\textbf{Step 2:} Without loss of generality, the second step uses the first biomarker to identify the values of $\delta_i$ for each patient. 

\begin{enumerate}[a)]
    \item First, we estimate the polynomial coefficients for each patient separately; we are guaranteed exact recovery of the coefficients by Assumption \ref{assum:enough_obs}. 
    \item Next we find the roots for each polynomial. If they are complex, consider their real part, and define $\xi_{i}$ to be the smallest root of the polynomial. At least one (real or complex) root is guaranteed to exist by the Fundamental Theorem of Algebra for every non-constant polynomial (Assumption \ref{assum:functionals}). Note that the choice of using the smallest root is arbitrary; what matters is that a consistent choice of root is selected for each patient's polynomials. 
    \item The goal of this step to learn a new polynomial for each patient which is translated to ensure that the root selected in step b) lies at $x=0$. 
    
    To do so, we first shift the observational time-steps by $\xi_i$, and we re-estimate the coefficients of each \emph{shifted} polynomial. 
    
    We make use of the fact that if $\xi_i$ is the smallest complex root of a polynomial $\kappa(x)$ then the polynomial $\kappa(x+\xi_i)$ has its smallest complex root at $0$. We can 
    recover the parameters of this polynomial exactly by shifting our observations and re-estimating the coefficients.
    
    This operation recovers the coefficients of every patient's polynomial in its \emph{canonical position} i.e. a translated polynomial whose the smallest root (or its real component) is at $x=0$. 
    
    This step can be viewed as a de-biasing step which allows us to re-estimate $\tilde{\theta}$ without while ignoring the effect
    that left-censorship has on parameter estimates. 
    
    \item We cluster the coefficients estimated in step c). 
    By construction, we know that $s_i = s_{i'}\iff \theta_i = \theta_{i'}$ which guarantees that clustering recovers the true-underlying subtype for each patient (up to a permutation over $K$ choices). 
    
    \item Finally we stratify patients by their subtype, and we define $\delta_i$ as the difference between their smallest root and the smallest value of $\xi_i$ among all other patients within that subtype. 
    
    By Assumption \ref{assum:delta0}, we know that for each subtype, there exists a patient for whom $\delta_i=0$, this reference patient will also be the one whose polynomial has the smallest root. We note here that without Assumption $\ref{assum:delta0}$, we would still have identification of $\delta_i$ up to a constant. 
    
    Therefore, by shifting each patient's smallest root by their reference patient's smallest root, we can recover the original time-shifts. 

\end{enumerate}
\textbf{Step 3:} Given the values of $\delta_1,\ldots,\delta_N$ from Step 2, we can now estimate the true values of the polynomial coefficients exactly in the noiseless setting via polynomial regression. 
\end{proof}